\documentclass[11pt]{wlscirep}
\usepackage[utf8]{inputenc}
\usepackage{float}
\usepackage[T1]{fontenc}
\usepackage[draft]{changes}
\usepackage{enumitem}
\usepackage{comment}
\newcommand{\citep}{\cite}
\usepackage{url}
\usepackage{xcolor}
\usepackage{seqsplit}
\usepackage{array}
\usepackage{tabularx}
\usepackage{multirow}
\usepackage{booktabs}
\usepackage{tabulary}
\usepackage{listings}
\usepackage{xcolor}
\usepackage{chngcntr}
\usepackage{caption}
\usepackage{hyperref}
\usepackage{cleveref}
\usepackage{xspace}
\DeclareCaptionType[fileext=lof,placement={!ht},within=none]{suppfigure}[Figure][List of Supplementary Figures]
\DeclareCaptionType[fileext=lot,placement={!ht},within=none]{supptable}[Table][List of Supplementary Tables]
\usepackage{hyperref}
\usepackage{nameref}
\definecolor{codegreen}{rgb}{0,0.6,0}
\definecolor{codegray}{rgb}{0.5,0.5,0.5}
\definecolor{codepurple}{rgb}{0.58,0,0.82}
\definecolor{backcolour}{rgb}{0.95,0.95,0.92}
\lstdefinestyle{mystyle}{
    backgroundcolor=\color{backcolour},
    commentstyle=\color{codegreen},
    keywordstyle=\color{magenta},
    stringstyle=\color{codepurple},
    basicstyle=\ttfamily\footnotesize,
    breakatwhitespace=false,         
    breaklines=true,                 
    keepspaces=true,                
    showspaces=false,                
    showstringspaces=false,
    showtabs=false,                  
    tabsize=2
}
\usepackage{xcolor}
\usepackage{tcolorbox}
\usepackage{caption}
\usepackage{wrapfig}
\usepackage{enumitem}
\usepackage{changepage}
\usepackage{subcaption}
\usepackage{colortbl} %
\definecolor{lightgray}{gray}{0.9}
\usepackage{threeparttable} %
\newtcolorbox{findingbox}{
  colback=white,
  colframe=blue!50!black,
  fonttitle=\bfseries,
  title={Finding},
  sharp corners,
  boxrule=1pt,
  left=0pt
}
\usepackage{xcolor}
\usepackage{fontawesome}
\DeclareRobustCommand{\huggingface}{\raisebox{-1.5pt}{\includegraphics[height=1.05em]{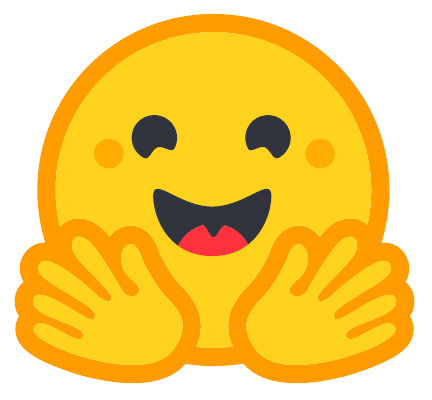}}\xspace}
\DeclareRobustCommand{\github}{\raisebox{-1.5pt}{\includegraphics[height=1.05em]{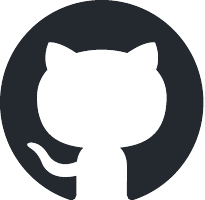}}\xspace}
\DeclareRobustCommand{\worldwideweb}{\raisebox{-1.5pt}{\includegraphics[height=1.05em]{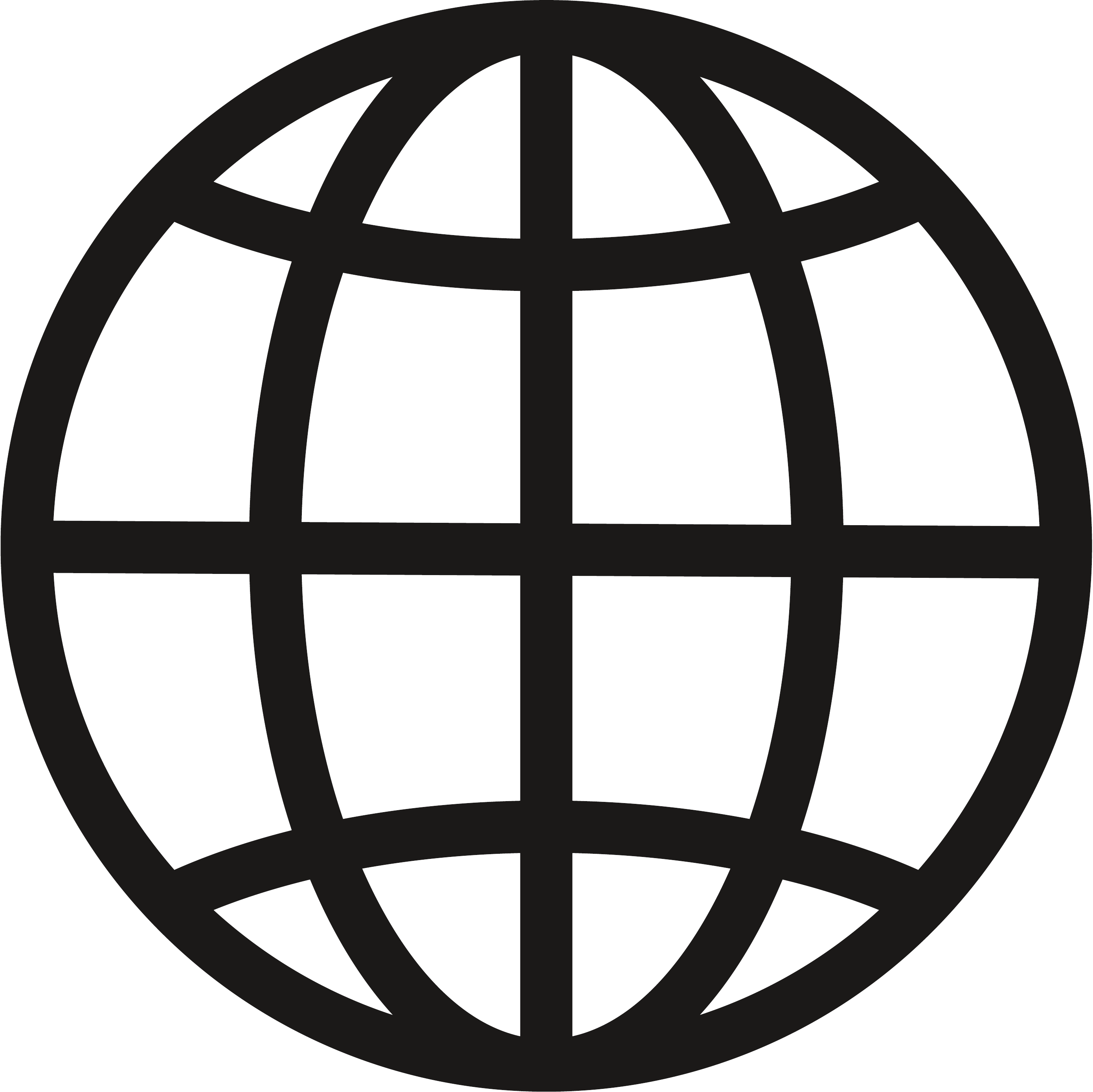}}\xspace}
\newcommand{\linkicon}[1]{\makebox[1.6em][l]{#1}}
\newcommand{\linklabel}[1]{\makebox[1.6cm][l]{\textbf{#1}}}

\title{
A General Model for Retinal Segmentation and Quantification
}

\author{RetSAM Team}
\affil{Monash University}
\affil{\linkicon{\worldwideweb{}} \linklabel{Website} \url{https://wzhjerry.github.io/RetSAM}}
\affil{\linkicon{\github{}} \linklabel{Code} \url{https://github.com/Wzhjerry/RetSAM}}
\affil{\linkicon{\huggingface{}} \linklabel{Model} \url{https://huggingface.co/JerryWzh/RetSAM_public}}

\begin{abstract}

Retinal imaging is fast, non-invasive, and widely available, offering quantifiable structural and vascular signals for ophthalmic and systemic health assessment. This accessibility creates an opportunity to study how quantitative retinal phenotypes relate to ocular and systemic diseases. However, such analyses remain difficult at scale due to the limited availability of public multi-label datasets and the lack of a unified segmentation-to-quantification pipeline. We present RetSAM, a general retinal segmentation and quantification framework for fundus imaging. It delivers robust multi-target segmentation and standardized biomarker extraction, supporting downstream ophthalmologic studies and oculomics correlation analyses. Trained on over 200,000 fundus images, RetSAM supports three task categories and segments five anatomical structures, four retinal phenotypic patterns, and more than 20 distinct lesion types. It converts these segmentation results into over 30 standardized biomarkers that capture structural morphology, vascular geometry, and degenerative changes. Trained with a multi-stage strategy using both private and public fundus data, RetSAM achieves superior segmentation performance on 17 public datasets. It improves on prior best methods by 3.9 percentage points in DSC on average, with up to 15 percentage points on challenging multi-task benchmarks, and generalizes well across diverse populations, imaging devices, and clinical settings. The resulting biomarkers enable systematic correlation analyses across major ophthalmic diseases, including diabetic retinopathy, age-related macular degeneration, glaucoma, and pathologic myopia. 
Together, RetSAM transforms fundus images into standardized, interpretable quantitative phenotypes, enabling large-scale ophthalmic research and translation.

\end{abstract}

\begin{document}
\maketitle



\section*{Introduction}

Retinal imaging is now routine in ophthalmic practice, enabled by rapid, non-invasive acquisition and the broad availability of fundus cameras in both specialist and community settings \cite{wong2001retinal,patton2006retinal,abramoff2010retinal,ting2017development}. Clinically, the retina offers direct visualization of neural tissue and microvasculature, allowing practitioners to assess a wide range of structural and vascular abnormalities from a single image. These signals are not only central to ophthalmic diagnosis, but also increasingly recognized as markers of broader physiology, motivating the emerging field of oculomics \cite{poplin2018prediction,wagner2020insights,zekavat2022deep,rim2020prediction,keane2018eye}. 

As large-scale retinal image collections continue to grow, the bottleneck is shifting from image acquisition to standardized quantitative characterization. Many clinically meaningful signals are still described qualitatively, and much of the literature remains at the level of detecting abnormalities rather than measuring them in a reproducible and comparable way across studies and populations \cite{welikala2014automated,trucco2015morphometric}. For both ophthalmic research and oculomics, scalable disease characterization requires standardized, interpretable, and reproducible biomarkers, allowing retinal signals to be measured consistently and linked to disease burden, progression, and risk at population scale \cite{zhou2022automorph}. 

Despite this clear need, delivering standardized retinal measurement at scale remains challenging for three reasons. First, training a single model with broad clinical coverage is limited by annotation resources: most public fundus datasets focus on a single task and use inconsistent definitions, making it difficult to learn one model that covers anatomy, lesions, and phenotypic patterns \cite{zhou2023foundation,litjens2017survey}. Second, segmentation is rarely paired with standardized quantification; measurement rules differ across papers and codebases, so biomarkers are hard to reproduce and compare across studies, limiting the translation of pixel-level predictions into patient-level insights \cite{knudtson2003revised,tatham2013relationship,perez2011vampire}. Third, real-world data distributions often show domain shifts: differences in cameras, acquisition protocols, and patient populations can affect both segmentation outputs and the resulting quantitative metrics \cite{wang2019patch,guan2021domain,lyu2022aadg}.

Here, we introduce RetSAM, a unified framework that combines multi-target fundus segmentation with standardized biomarker extraction to support ophthalmic disease correlation studies and discovery-oriented oculomics. RetSAM follows three principles: unified coverage across retinal targets, reproducible biomarker extraction, and clinical usability. The framework supports a single pipeline that covers a unified target set spanning anatomical structures, lesions, and phenotypic patterns, enabling robust multi-target segmentation from a single image. Building on these outputs, RetSAM produces standardized and interpretable quantitative biomarkers with consistent measurement definitions and structured, report-ready summaries. This makes the resulting measurements comparable across studies and ready for downstream ophthalmologic research, oculomics analyses, and clinical reporting and workflow integration studies. 

Technically, RetSAM is built around a unified multi-target segmentation core and a modular quantification engine. To handle fragmented annotations, RetSAM maps task-specific labels into a shared label space with consistent definitions across retinal targets, enabling joint learning within a single model. Training follows a step-wise curriculum, using supervision aggregated from both public and private datasets to improve robustness across acquisition settings. A modular quantification engine then applies standardized measurement rules to the segmentation outputs and converts task-specific segmentations into clinically relevant measurements, such as vessel fractal dimension \cite{knudtson2003revised,lim2009retinal}, cup-to-disc ratio \cite{tatham2013relationship,gu2019net}, and lesion size and count. 

We provide evidence for RetSAM across technical benchmarking, scientific validation, and clinical translation. On 17 public fundus benchmarks spanning diverse segmentation targets, RetSAM achieves competitive and consistent performance, providing a reliable foundation for downstream quantification. Based on the resulting biomarkers, RetSAM enables correlation analyses on public datasets for diabetic retinopathy, age-related macular degeneration, glaucoma, and pathologic myopia. RetSAM also supports discovery-oriented oculomics analyses of eye--brain, eye--heart, and eye--kidney relationships by providing harmonized quantitative phenotypes suitable for population-scale association studies. Finally, RetSAM is designed for clinical translation through biomarker-supported outputs that can be evaluated in a multinational reader study \cite{milea2020artificial,sayres2019using} and a prospective randomized controlled trial \cite{lin2019diagnostic} to assess effects on clinician confidence, workflow efficiency, and diagnostic accuracy.

To summarize, the core contributions of RetSAM are:
\begin{itemize}
    \item We introduce RetSAM, a unified fundus segmentation-to-quantification framework that produces consistent multi-target masks and standardized retinal biomarkers within a single pipeline, enabling reproducible phenotyping across imaging domains, devices, and populations. 
    \item We demonstrate the effectiveness of RetSAM through comprehensive benchmarking against representative baselines, achieving superior performance across 17 public downstream segmentation datasets and providing a reliable foundation for large-scale quantitative analysis. 
    \item We release RetSAM as an open-source toolkit, including trained models and the end-to-end segmentation-to-quantification pipeline, to support reproducible research and broader validation across datasets and studies. 
\end{itemize}

\section*{Results}

\subsection*{Overview of RetSAM}

\begin{figure*}[] 
\centering
\includegraphics[width=\textwidth]{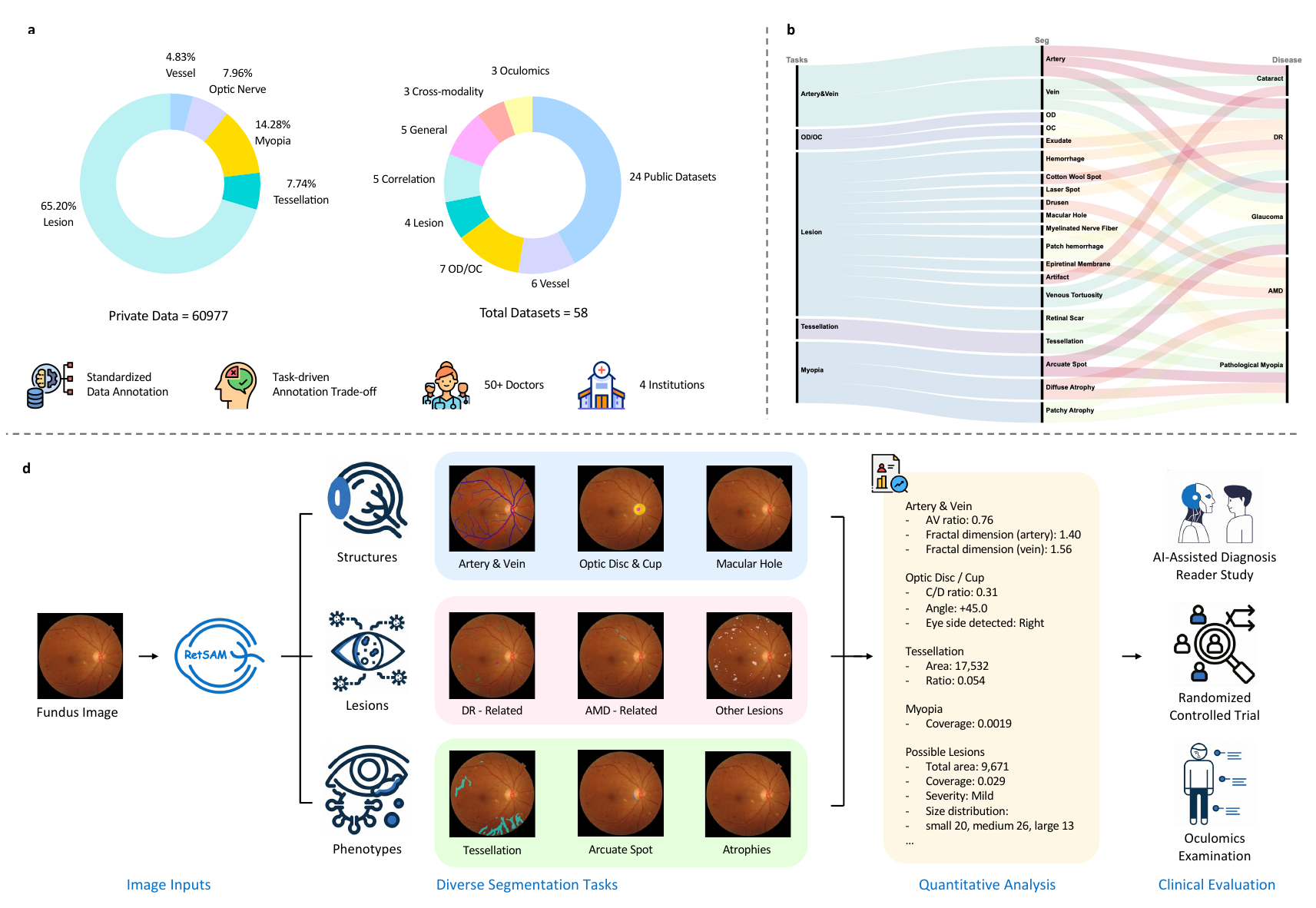}
\captionsetup{justification=justified, singlelinecheck=false}
\caption{
\textbf{Overview of the RetSAM workflow.} \textbf{(a) Data curation and clinical integration:} RetSAM is trained and evaluated on a comprehensive cohort of public and private fundus datasets, utilizing a unified annotation protocol that combines labeled and unlabeled data. Over 50 ophthalmologists from four clinical institutions contributed to the project, participating in data curation, reader studies, and a randomized controlled trial. \textbf{(b) Task-disease alignment:} RetSAM’s segmentation tasks are mapped to clinically relevant ophthalmologic disease manifestations, bridging structural and lesion patterns with diagnostic categories used in routine practice. \textbf{(c) Segmentation, quantification, and validation:} From a single fundus image, RetSAM jointly segments retinal structures and lesion phenotypes to derive over 30 quantitative biomarkers. These biomarkers facilitate diagnosis, disease monitoring, and oculomic association analysis. The system underwent rigorous validation via a multinational reader study and a prospective randomized controlled trial, demonstrating enhanced diagnostic confidence and clinical decision-making efficiency.
}
\label{fig:main_figure}
\end{figure*}

RetSAM is a unified framework for scalable segmentation and quantitative characterization from fundus images. It integrates multi-target segmentation with a downstream quantification pipeline, as shown in Figure~\ref{fig:main_figure}. RetSAM organizes its segmentation outputs into three groups: anatomical structures, lesions, and phenotypic patterns. This separation distinguishes (i) structures present in nearly all eyes, (ii) localized signs of disease, and (iii) non-lesion patterns that appear in pathological conditions and reflect broader changes in retinal appearance.

Given an input fundus image, RetSAM produces multi-target segmentation predictions across these three groups and derives a standardized set of quantitative biomarkers from the predicted masks for downstream analysis. The system exports both visual overlays and structured biomarker outputs for cohort-scale studies. A detailed description of the segmentation tasks, output definitions, and the complete biomarker set is provided in Table~\ref{tab:segmentation_tasks}, while implementation details are described in the Methods section.

\subsection*{Evaluation Protocols}

We evaluate RetSAM from three perspectives: technical validation, clinical application and discovery, and human--AI collaboration. The data distribution is visualized in Figure~\ref{fig:main_figure}(a), and full details are provided in the Methods section.

\paragraph{Technical Validation.} We conduct comprehensive benchmarking by reporting task-specific metrics and comparing against representative baseline methods in three categories. These include SAM-based approaches such as MedSAM \cite{ma2024segment}, MedSAM2 \cite{zhu2024medical}, Medical-SAM-Adapter \cite{wu2025medical}, SAM2 \cite{ravi2024sam}, and SAM3 \cite{carion2025sam}, fully supervised models represented by SAM2-UNet \cite{xiong2024sam2}, and domain-specific foundation models such as RetFound \cite{zhou2023foundation}. The primary datasets used for these evaluations are summarized in Table~\ref{tab:downstream_datasets}. First, to assess fine-grained task-specific precision, we evaluate RetSAM on 17 publicly available datasets \cite{fraz2012ensemble,staal2004ridge,jin2022fives,budai2013robust,zhang2016robust,hoover2000locating,carmona2008identification,sivaswamy2014drishti,bajwa2020g1020,wu2023gamma,zhang2010origa,kovalyk2022papila,orlando2020refuge,li2019diagnostic,decenciere2013teleophta,zhou2020benchmark,porwal2018indian}. These datasets cover the framework's primary output groups, including retinal vessels and optic disc/cup (OD/OC) representing anatomical structures, along with various pathological lesions. Second, we examine the model's general segmentation capability using one public benchmark \cite{lepetit2024maples} and four datasets that we re-annotated to ensure label consistency. To assess the effectiveness of our unified annotation space and the teacher-based multi-stage training strategy, we utilize the private datasets used for multi-stage training listed in Table~\ref{tab:private_datasets}, specifically to quantify the benefits of multi-target joint training. Finally, we assess robustness to domain shifts via cross-modality evaluation on two public datasets \cite{ding2020weakly,li2024octa} and one private dataset spanning two distinct imaging modalities, as listed in Table~\ref{tab:cross_modal_datasets}.

\subsection*{Technical Validation}

\subsubsection*{Public Benchmark Evaluation}

We evaluate RetSAM on 17 public datasets spanning three tasks: retinal vessel segmentation, OD/OC, and lesion segmentation. These datasets differ in clinical centers, imaging protocols, and camera types. Despite these variations, RetSAM generalizes well across the benchmarks. For vessel segmentation, RetSAM achieves competitive DSC and JAC scores compared with other methods, as shown in Table~\ref{tab:vessel_performance_all}. Beyond pixel-level overlap, RetSAM also preserves vascular topology. It attains an average clDice of 34.7\%, exceeding the strong baseline SAM2-UNet by over 14\%, which is consistent with fewer fragmented predictions and improved connectivity for thin capillaries.

For OD segmentation (Table~\ref{tab:od_performance}), RetSAM performs comparably to leading approaches. While it is not the top performer on every dataset, the gap is small, as this task is mature and improvements are typically incremental. For OC segmentation, the fine-tuned RetSAM achieves leading results on most datasets (Table~\ref{tab:oc_performance}). We observe lower performance under the linear inference protocol, likely due to differences in annotation protocols, particularly cup boundary definitions between our pre-training data and certain public benchmarks. Fine-tuning largely mitigates this discrepancy and improves alignment with the target annotation style.

For lesion segmentation (Tables~\ref{tab:lesion_dsc_transposed} and~\ref{tab:lesion_precision_transposed}), RetSAM shows strong adaptability across datasets. For hard exudates and soft exudates, RetSAM leads on benchmarks such as IDRiD and DDR. We observe performance variation on FGADR, likely due to its annotation format differing from our private data. Despite this, RetSAM adapts to the dataset-specific labeling conventions and remains competitive. For microaneurysms, RetSAM reduces false negatives and produces more precise segmentations. On some datasets, RetSAM yields slightly lower DSC than certain baselines but substantially higher precision, suggesting reduced over-segmentation and higher-specificity outputs that are more suitable for clinical quantification.

\subsubsection*{General Segmentation Ability}

We further assess RetSAM's ability to perform simultaneous multi-target segmentation. Unlike the previous section, this evaluation focuses on samples that require concurrent predictions for multiple anatomical targets. To build a rigorous benchmark, we use one fully annotated public dataset and re-annotate four additional datasets by adding missing labels for vessels, OD/OC, and lesions to ensure comprehensive coverage. The quantitative results in Tables~\ref{tab:general_performance_part1} and~\ref{tab:general_performance_maples} summarize performance in this setting. RetSAM achieves leading results across evaluated tasks and outperforms competing methods. These results suggest that the model learns shared representations across targets and maintains balanced performance across tasks. In contrast to typical multi-task settings where easier tasks can dominate harder ones, RetSAM maintains accuracy across classes regardless of which targets are present. On the IDRiD benchmark, RetSAM achieves an average gain of 9.8\% across structural tasks over the second-best method. The largest improvement is observed for vessel segmentation, where RetSAM increases DSC by 18.7\% compared with SAM2-UNet. Together, these results support RetSAM as a unified framework for multi-target fundus analysis.

\subsubsection*{Evaluation on Pseudo-Generated Data}

To assess our multi-stage strategy, we benchmark RetSAM against the single-task expert models used for pseudo-label generation. Table~\ref{tab:ablation_study} compares these baselines with RetSAM under both linear inference and fine-tuning protocols. RetSAM consistently outperforms the single-task experts across evaluated metrics. Notably, even with a frozen backbone, RetSAM exceeds the baselines that produced its supervision labels. For example, in OC segmentation, linear inference shows a 7.5\% improvement over the expert baseline, and we observe consistent gains in artery/vein segmentation. These results suggest that RetSAM does not simply reproduce pseudo-labels, but instead learns representations that generalize beyond individual single-task teachers. A likely contributor is multi-target learning, where jointly modeling diverse targets encourages the encoder to capture shared fundus characteristics and reduces sensitivity to target-specific noise. Fine-tuning further improves performance, indicating that the multi-target representation provides a strong initialization for task-specific adaptation.

\subsubsection*{Data and Training Efficiency}

\begin{figure*}[] 
\centering
\includegraphics[width=\textwidth]{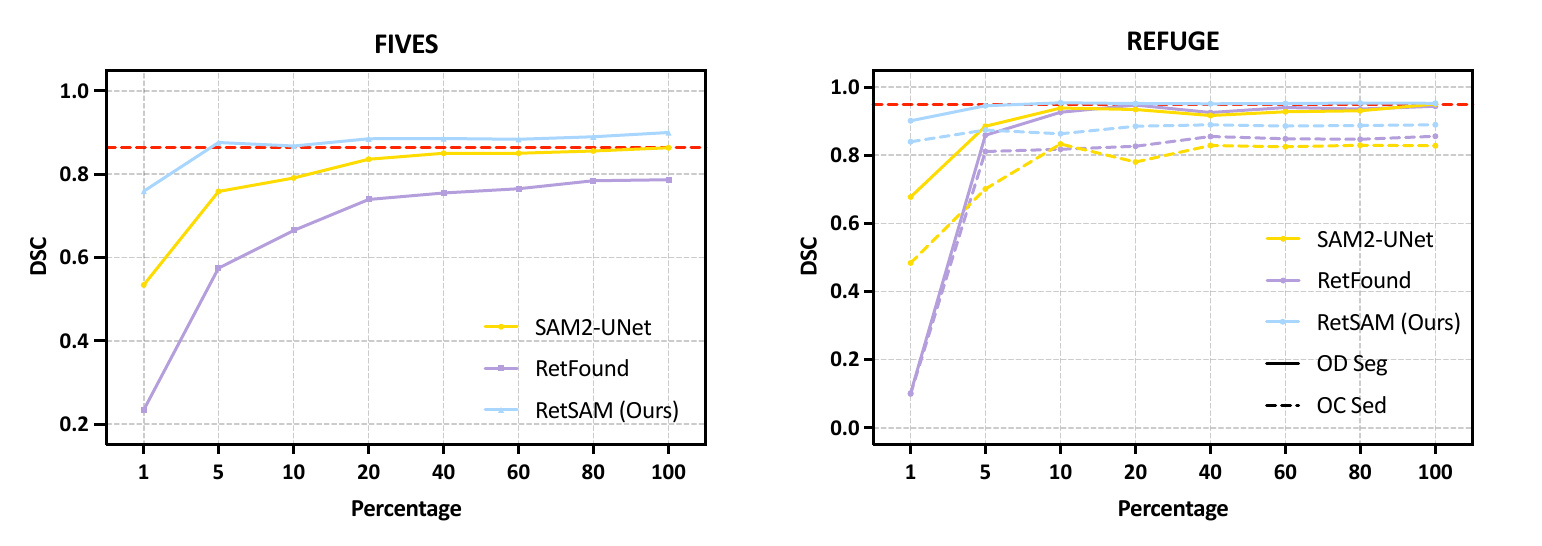}
\captionsetup{justification=justified, singlelinecheck=false}
\caption{
Comparison of segmentation performance under limited supervision. We evaluate RetSAM against RetFound and SAM2-UNet on the FIVES and REFUGE datasets using different ratios of training data. The red dashed line indicates the specific data fraction required by RetSAM to match the peak performance attained by the best competing method using full data. 
}
\label{fig:data_efficiency}
\end{figure*}

To evaluate performance in label-scarce regimes, we conduct experiments on two representative datasets covering vessel segmentation (FIVES) and OD/OC segmentation (REFUGE). We simulate varying data availability by randomly sampling fractions of the training set: \{1\%, 5\%, 10\%, 20\%, 40\%, 60\%, 80\%, 100\%\}. Figure~\ref{fig:data_efficiency} shows performance trajectories for RetSAM compared with SAM2-UNet and RetFound. RetSAM matches or exceeds baseline performance using only 5\% of the annotated data and remains strong even at the 1\% regime. This suggests that the multi-target pre-trained encoder provides transferable features, reducing the reliance on large amounts of task-specific supervision. As more target-domain data are added, gains are primarily driven by aligning these features with dataset-specific characteristics and annotation styles. Overall, RetSAM exhibits strong sample efficiency, supporting efficient adaptation to downstream tasks and new domains.

\subsubsection*{Cross-Modality Generalization}

We extend evaluation to ultra-widefield fundus imaging (UWF) and optical coherence tomography angiography (OCTA). Because UWF shares core photometric characteristics and anatomical features with standard fundus photography, despite differences in field of view and geometric distortion, we hypothesize that RetSAM can transfer effectively. To test transferability under limited supervision, we use a few-shot protocol where only 5\% of the target dataset is available for training. For a fair comparison, all competing baselines are trained from scratch using the same 5\% subset. Quantitative results are reported in Table~\ref{tab:cross_modal_performance}.

For UWF, RetSAM is robust to geometric deformation. On the private UWF dataset, RetSAM under linear inference surpasses SAM2-UNet that is fine-tuned on the 5\% subset, suggesting that the learned representations transfer across fields of view. After fine-tuning on 5\% data, RetSAM further improves accuracy and converges faster than the baselines under the same data constraints.

In contrast, OCTA introduces a larger domain shift due to its depth-resolved appearance and lack of color information, making linear inference ineffective. We therefore fine-tune all models using the full OCTA training set. Despite this modality gap, RetSAM provides a stronger initialization than standard weights. RetSAM consistently outperforms SAM2-UNet trained on the same full dataset for both artery and vein segmentation. These results indicate that even when domain shifts preclude zero-shot transfer, RetSAM's pre-trained features support better optimization and stronger final performance than general-purpose encoders.

\subsection*{Qualitative Comparisons}

\begin{figure*}[] 
\centering
\includegraphics[width=\textwidth]{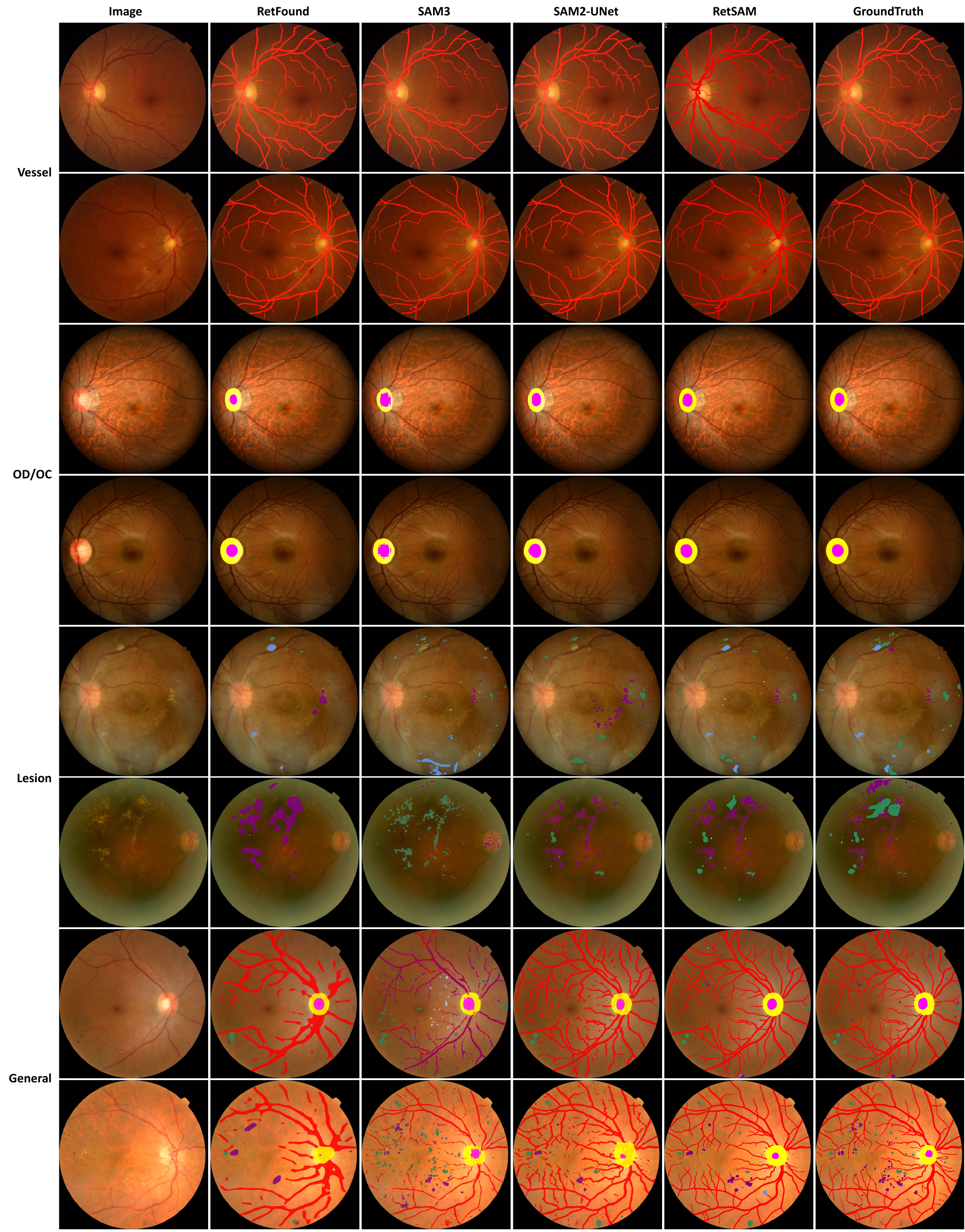}
\captionsetup{justification=justified, singlelinecheck=false}
\caption{
Qualitative segmentation comparison of RetSAM against RetFound, SAM3, and SAM2-UNet across four downstream tasks. The predicted segmentation masks are overlaid on the original images with distinct colors. 
}
\label{fig:qualitative_comparisons}
\end{figure*}

We compare RetSAM against RetFound, SAM3, and SAM2-UNet on FIVES, GAMMA, DDR, and MAPLES-DR. Figure~\ref{fig:qualitative_comparisons} shows that RetSAM consistently produces higher-quality segmentations. Baselines occasionally yield fragmented regions or boundary artifacts, whereas RetSAM generates smoother and more continuous contours that adhere closely to anatomical structures. These qualitative trends are consistent with the quantitative results, supporting improved edge fidelity and shape consistency across tasks.

\begin{figure*}[] 
\centering
\includegraphics[width=\textwidth]{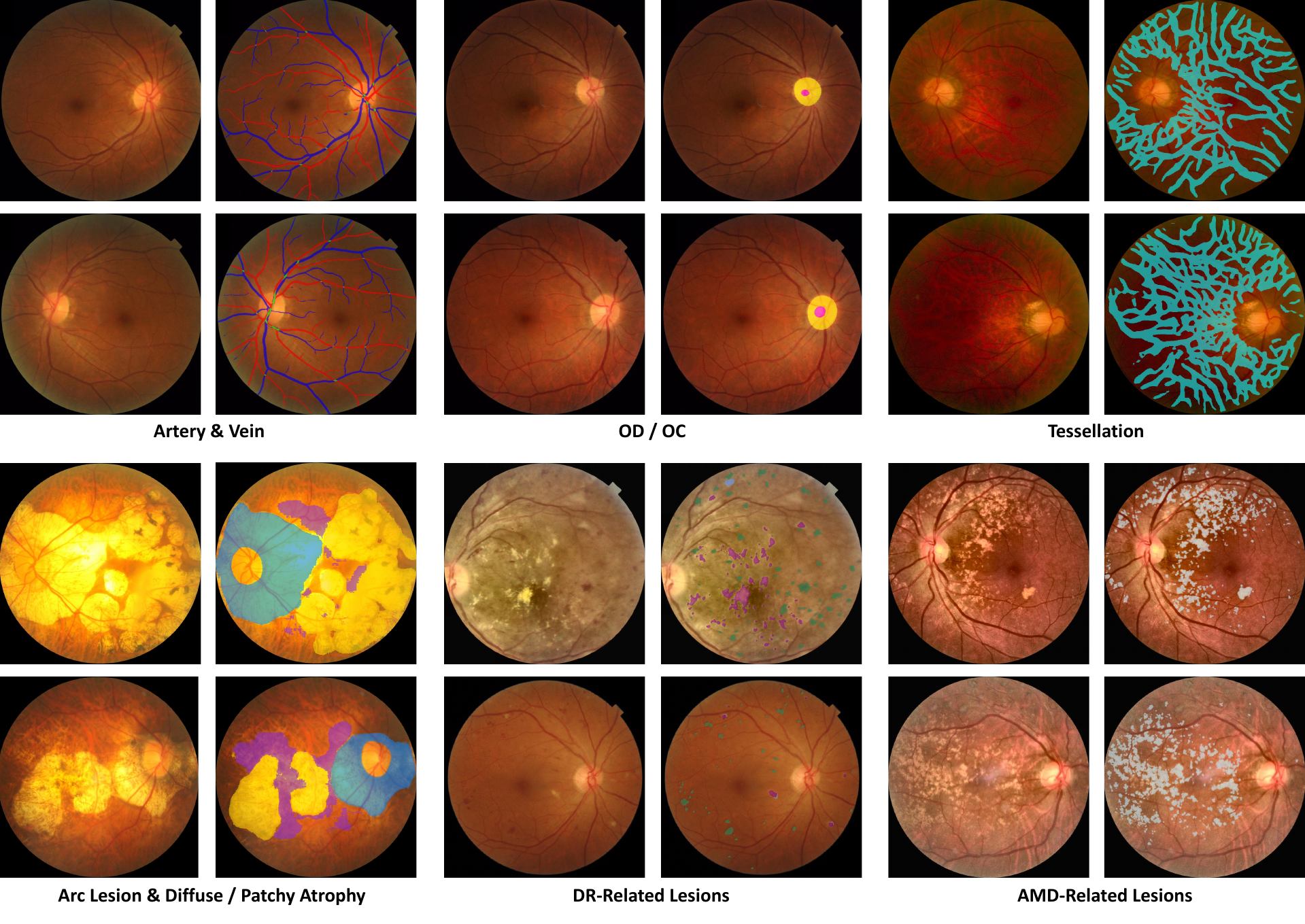}
\captionsetup{justification=justified, singlelinecheck=false}
\caption{
Qualitative demonstration of RetSAM's comprehensive segmentation capabilities. Segmentation predictions are displayed as overlays with distinct colors. The figure highlights the diverse range of categories the model can identify, spanning from anatomical structures to pathological lesions and fundus features. 
}
\label{fig:qualitative_examples_all}
\end{figure*}

We further illustrate RetSAM's coverage across anatomical and pathological targets. As shown in Figure~\ref{fig:qualitative_examples_all}, the model segments fine-grained structures such as arteries, veins, optic cup, and optic disc, as well as phenotypic patterns including tessellated fundus, arcuate spots, and diffuse or patchy atrophy. The model also provides precise segmentations for lesions associated with diabetic retinopathy and age-related macular degeneration, highlighting its ability to operate as a unified framework across categories.

\begin{figure*}[] 
\centering
\includegraphics[width=\textwidth]{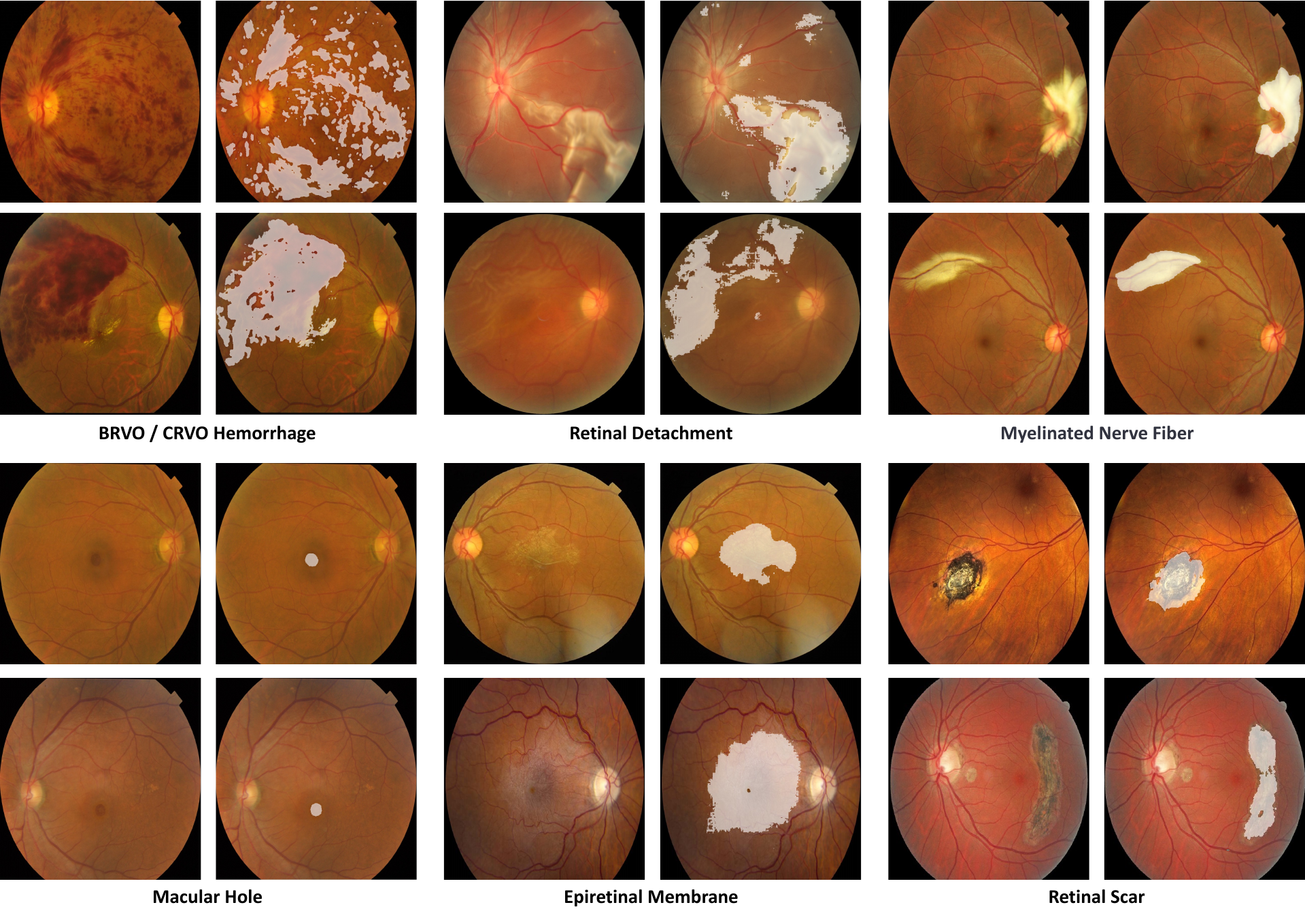}
\captionsetup{justification=justified, singlelinecheck=false}
\caption{
Qualitative segmentation results for retinal lesions beyond common DR and AMD-related categories. Segmentation predictions are displayed as white overlays. These examples demonstrate the model's effective coverage of lesion classes which are frequently absent from general-purpose public benchmarks. 
}
\label{fig:other_lesions}
\end{figure*}

To assess robustness beyond common lesion types, Figure~\ref{fig:other_lesions} presents results for additional fundus lesions. RetSAM identifies these targets reliably, suggesting that its learned representations extend beyond the most frequent disease categories encountered during training and remain effective in more diverse clinical scenarios.

\begin{figure*}[] 
\centering
\includegraphics[width=\textwidth]{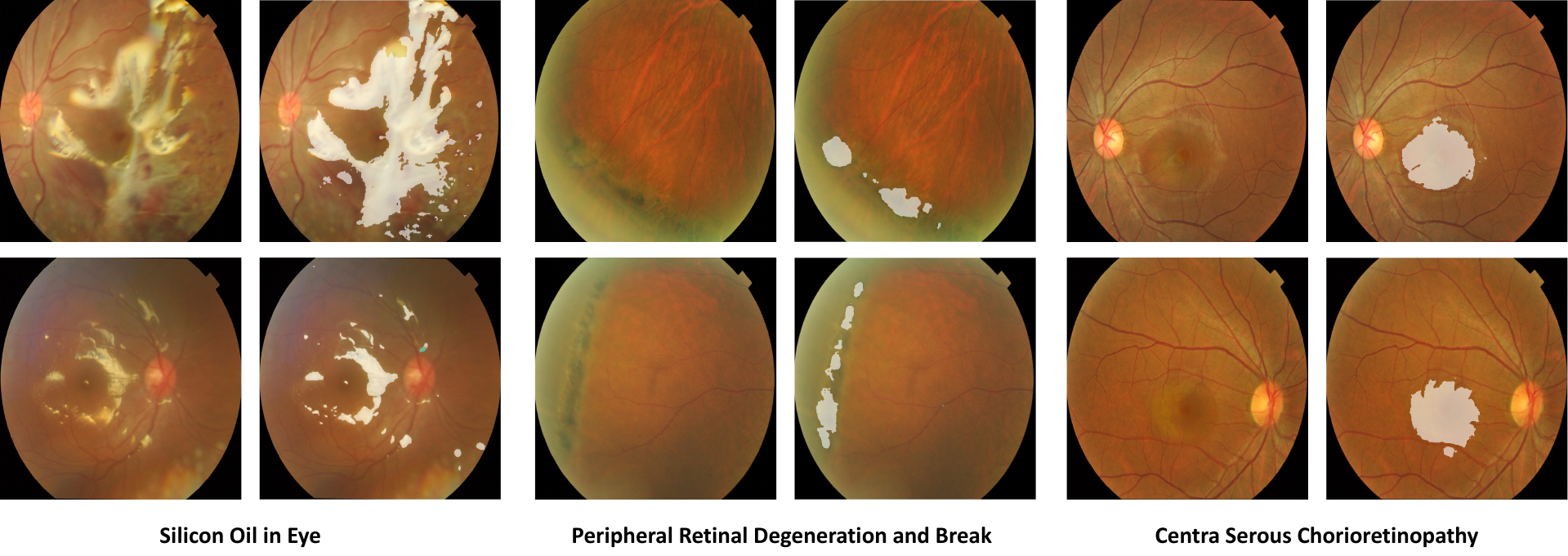}
\captionsetup{justification=justified, singlelinecheck=false}
\caption{
Qualitative representations of OOD lesion segmentation. Segmentation predictions are displayed as white overlays. The examples show specific pathologies that were entirely absent from the training dataset. Despite this, RetSAM successfully captures these unseen areas, demonstrating that it has learned a generalized concept of retinal abnormalities beyond specific named classes. 
}
\label{fig:ood_lesions}
\end{figure*}

We also evaluate out-of-distribution lesions that were explicitly excluded from the training set \cite{cen2021automatic}, as shown in Figure~\ref{fig:ood_lesions}. Although these specific classes are unseen, the model can localize anomalous regions. This behavior is supported by the inclusion of a generic lesion category during training, supervised with an aggregated set of diverse lesion types, which encourages the model to learn a general representation of abnormal retinal appearance. As a result, RetSAM can distinguish abnormal textures from background tissue and segment out-of-distribution lesions based on visual deviation, even without access to their semantic labels.

\begin{figure*}[] 
\centering
\includegraphics[width=0.75\textwidth]{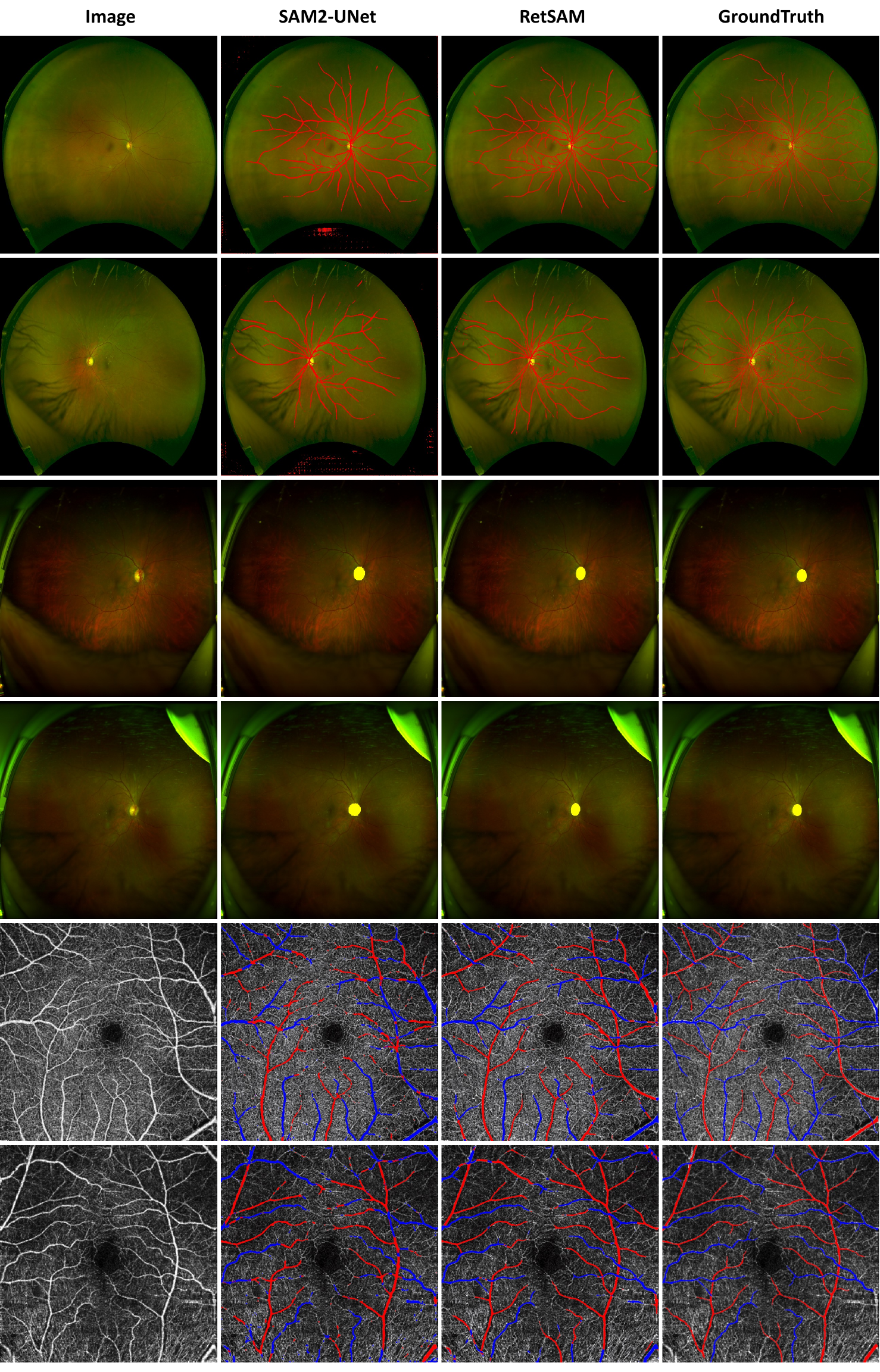}
\captionsetup{justification=justified, singlelinecheck=false}
\caption{
Qualitative cross-modality segmentation comparison between RetSAM and SAM2-UNet. Columns show results for vessel and OD segmentation on UWF images, and artery \& vein segmentation on OCTA images. RetSAM demonstrates superior adaptability and structural preservation across these diverse modalities compared to SAM2-UNet. 
}
\label{fig:qualitative_cross_modality}
\end{figure*}

Finally, we qualitatively assess cross-modality scenarios on UWF and OCTA compared with SAM2-UNet, as shown in Figure~\ref{fig:qualitative_cross_modality}. RetSAM adapts better across both modalities. In OCTA, despite the absence of color information, the model separates arteries and veins more clearly and produces cleaner vessel maps than the baseline. In UWF, RetSAM delineates the optic disc and major vascular arcades while handling geometric distortion. We note a limitation for UWF vessel segmentation. Although RetSAM substantially improves over the baseline, it can omit fine peripheral capillary networks. This is likely related to the extreme native resolution of UWF images relative to the fixed model input size. Resizing reduces high-frequency spatial detail, making some peripheral capillaries difficult to resolve after downsampling. 

\section*{Discussion}

Despite the progress of deep learning in medical imaging, its application in ophthalmology remains fragmented. While current models perform well on specific segmentation tasks, they rarely address subsequent automated quantification or comprehensive clinical analysis. This limitation, coupled with the need for separate tools, restricts diagnostic efficiency. Consequently, a unified foundation model is essential to accurately segment diverse retinal structures and lesions, enabling holistic and quantifiable analysis.

In this study, we introduce RetSAM, a unified foundation model developed on a large collection of private and public datasets. RetSAM demonstrates robust segmentation capabilities, accurately identifying diverse anatomical structures and lesions within the fundus, and achieving leading performance in downstream task validation. Uniquely, it integrates an automated analysis pipeline that outputs 30 morphological metrics, directly transforming raw imaging data into quantitative insights for clinical research. In terms of core performance, RetSAM exhibits strong generalization capabilities, significantly outperforming competing approaches even in scenarios with limited annotated samples. This efficiency provides a robust solution for adapting to diverse and unseen retinal conditions.

While generalist models like MedSAM show potential, their performance in ophthalmology is often limited. This is primarily because they lack sufficient training on domain-specific data and do not cover the specific requirements of diverse downstream tasks. RetSAM addresses this issue by incorporating a large collection of both private and public datasets, resulting in robust performance across various retinal applications. Additionally, a key limitation of current generalist models is the lack of automated quantification. The need for both higher precision and quantitative analysis supports the development of a dedicated ophthalmic foundation model. RetSAM fills this gap, offering not only accurate segmentation, but also the quantitative insights required for clinical research.

Beyond functioning as a standalone segmentation model, RetSAM serves as a core perception tool within ophthalmic AI agent systems. In this capacity, it translates complex image features such as vessel morphology, lesion areas, and cup-to-disc ratios into structured quantitative metrics. This conversion capability is essential for integrating with Large Language Models (LLMs). While LLMs excel at logical reasoning and text processing, they lack the ability to directly and precisely quantify raw visual data. By providing accurate, text-based oculomic indicators, RetSAM acts as a bridge, enabling LLMs to perform grounded pathological analysis, report generation, and risk prediction. This integration effectively combines precise visual perception with reasoning capabilities.

Despite its promising capabilities, RetSAM has certain limitations. Currently, the model is primarily optimized for color fundus photography. While we demonstrate adaptability to UWF and OCTA via fine-tuning, its zero-shot performance on 3D volumetric modalities requires further study. Additionally, while RetSAM covers a broad spectrum of pathologies, the segmentation of extremely rare lesions or atypical presentations may still present challenges due to data scarcity. Finally, although the downstream tasks encompass the majority of retinal structures and lesions, the task settings remain relatively fixed. This predefined configuration ensures standardization but may constrain the model's adaptability to open-ended clinical inquiries compared to interactive prompt-based approaches.

In conclusion, RetSAM demonstrates the effectiveness of foundation model strategies for precise ophthalmic segmentation. By combining robust multi-target segmentation with standardized quantification, it offers a practical framework for both clinical workflows and large-scale research. Ultimately, RetSAM provides the quantitative foundation necessary for efficient automated screening and data-driven ophthalmic analysis.

\section*{Contributed Authors}
Zhonghua Wang$^{1}$, Lie Ju$^{1,2,3}$, Sijia Li$^{1}$, Wei Feng$^{1}$, Sijin Zhou$^{1}$, Ming Hu$^{1}$, Jianhao Xiong$^{5}$, Xiaoying Tang$^{6}$, Yifan Peng$^{7}$, Mingquan Lin$^{8}$, Yaodong Ding$^{9}$, Yong Zeng$^{9}$, Wenbin Wei$^{4}$, Li Dong$^{4}$, Zongyuan Ge$^{1}$

\medskip
\noindent $^{1}$AIM for Health Lab, Faculty of Information Technology, Monash University, Melbourne, Australia\\
$^{2}$Institute of Ophthalmology, University College London, London, UK\\
$^{3}$NIHR Biomedical Research Centre at Moorfields Eye Hospital NHS Foundation Trust, London, UK\\
$^{4}$Beijing Tongren Eye Center, Beijing Tongren Hospital, Capital Medical University, Beijing, China\\
$^{5}$Airdoc LLC., Beijing, China\\
$^{6}$Department of Electronic and Electrical Engineering, Southern University of Science and Technology, Shenzhen, China\\
$^{7}$Population Health Sciences, Weill Cornell Medicine, New York, NY, USA\\
$^{8}$Division of Computational Health Sciences, Department of Surgery, University of Minnesota, Minneapolis, MN, USA\\
$^{9}$Department of Cardiology, Beijing Anzhen Hospital, Capital Medical University, Beijing, China

\clearpage
\bibliography{ref}
\clearpage

\clearpage 
\newpage

\clearpage 
\newpage

\clearpage 
\newpage

\clearpage 
\newpage

\clearpage 
\newpage

\clearpage 
\newpage

\clearpage 
\newpage

\begin{table*}[t]
    \centering
    \caption{Summary of Segmentation Tasks and Output Definitions for RetSAM. }
    \label{tab:segmentation_tasks}
    \small
    \renewcommand{\arraystretch}{1.3} 
    
    \begin{tabularx}{\textwidth}{
        >{\raggedright\arraybackslash}p{0.10\textwidth} 
        >{\raggedright\arraybackslash}p{0.12\textwidth} 
        >{\raggedright\arraybackslash}p{0.18\textwidth} @{\hspace{1em}} 
        X
    }
        \toprule
        \textbf{Category} & \textbf{Task} & \textbf{Specific Target} & \textbf{Description} \\
        \midrule
        
        \multirow[t]{4}{=}{Anatomical Structures} 
        & \multirow[t]{2}{=}{Vessel} 
        & Artery & Vessels that transport oxygenated blood to the retina; typically lighter in color and narrower than veins. \\
        & & Vein & Vessels that drain deoxygenated blood; typically darker and wider compared to arteries. \\
        \cmidrule(r){2-4} 
        
        & \multirow[t]{2}{=}{Optic Nerve} 
        & Optic Disc & The exit point of retinal ganglion cell axons and entry point for blood vessels; corresponds to the physiological blind spot. \\
        & & Optic Cup & The central depression within the optic disc; its enlargement relative to the disc is a key indicator of glaucoma. \\
        \midrule
        
        \multirow[t]{4}{=}{Phenotypes} 
        & Tessellation 
        & Tessellation 
        & Visibility of large choroidal vessels due to the thinning or hypopigmentation of the Retinal Pigment Epithelium (RPE). \\
        \cmidrule(r){2-4}
        
        & \multirow[t]{3}{=}{Myopic Features} 
        & Peripapillary Atrophy & Crescent-shaped or irregular zones of chorioretinal atrophy adjacent to the optic disc, characteristic of myopic eyes. \\
        & & Diffuse Atrophy & Ill-defined, yellowish lesions indicating partial loss of choroidal tissue and the RPE layer. \\
        & & Patchy Atrophy & Well-defined, whitish-gray areas representing complete atrophy of the RPE and choriocapillaris. \\
        \midrule
        
        \multirow[t]{11}{=}{Lesions} 
        
        & \multirow[t]{4}{=}{DR Lesions} 
        & Hemorrhage & Rupture of blood vessels causing leakage; manifests as dot, blot, or flame-shaped blood spots. \\
        & & Exudates & Lipid and lipoprotein residues leaking from damaged capillaries; appear as bright, reflective yellow spots. \\
        & & Cotton Wool Spots & Fluffy white patches caused by blockage of axoplasmic transport in the nerve fiber layer due to localized ischemia. \\
        & & Laser Spot & Round atrophic or pigmented scars resulting from panretinal photocoagulation treatment. \\
        \cmidrule(r){2-4}
        
        & \multirow[t]{2}{=}{AMD Lesions} 
        & Drusen & Extracellular accumulation of lipids and proteins between the RPE and Bruch's membrane; a hallmark sign of AMD. \\
        & & Patchy Hemorrhage & Large areas of subretinal or sub-RPE bleeding, often associated with choroidal neovascularization. \\
        \cmidrule(r){2-4}
        
        & \multirow[t]{4}{=}{Other Lesions} 
        & Epiretinal Membrane & Fibrocellular tissue proliferation on the inner retinal surface that can cause macular distortion. \\
        & & Macular Hole & A full-thickness defect of the retinal tissue involving the anatomic fovea, severely affecting central vision. \\
        & & Artifacts & Non-pathological noise in the image, such as corneal reflections, eyelash shadows, or dust on the lens. \\
        & & Retinal Scar & Fibrous tissue formation resulting from prior trauma, inflammation, or healed pathological lesions. \\
        \cmidrule(r){2-4}
        
        & Possible Lesions 
        & Other Possible Lesions 
        & Additional fundus lesions not categorized under the primary DR or AMD tasks due to limited training samples. This class encompasses Edema, Arteriovenous nicking, Venous beading, Vascular sheathing, Pigmentary changes, Fibrous proliferation, Vitreous degeneration, and others. \\
        
        \bottomrule
    \end{tabularx}
\end{table*}

\clearpage
\newpage

\begin{table*}[t]
    \centering
    \caption{Summary of public datasets used for knowledge distillation and pseudo-label generation. }
    \label{tab:public_datasets}
    \small
    \renewcommand{\arraystretch}{1.3} 
    
    \begin{tabularx}{\textwidth}{
        >{\raggedright\arraybackslash}p{0.16\textwidth} 
        >{\raggedright\arraybackslash}p{0.12\textwidth} 
        >{\raggedright\arraybackslash}p{0.08\textwidth} 
        X
    }
        \toprule
        \textbf{Dataset} & \textbf{Country} & \textbf{Images} & \textbf{Link to Dataset} \\
        \midrule
        
        Acrima 
        & Spain & 705
        & https://figshare.com/s/c2d31f850af14c5b5232 \\
        
        BEH 
        & Bangladesh & 634
        & https://drive.google.com/file/d/1YdZm-sioiAbTdBRy4oej1q6tZL8Baft7/view \\
        
        DDR 
        & China & 13,673
        & https://github.com/nkicsl/DDR-dataset \\
        
        DeepDRiD 
        & China & 2000
        & https://isbi.deepdr.org/data.html \\

        DiaRetDB0 
        & Finland & 130
        & http://www.it.lut.fi/project/imageret/diaretdb0/ \\

        DiaRetDB1 
        & Finland & 89
        & http://www.it.lut.fi/project/imageret/diaretdb1/ \\
        
        DR1 
        & Brazil & 1077
        & https://doi.org/10.6084/m9.figshare.953671 \\
        
        DR2 
        & Brazil & 520
        & https://doi.org/10.6084/m9.figshare.953671 \\
        
        EyePACS 
        & USA & 88,702
        & https://www.kaggle.com/c/diabetic-retinopathy-detection \\
        
        FIRE 
        & Greece & 268
        & https://www.ics.forth.gr/cvrl/fire/ \\
        
        HEI-MED
        & USA & 169
        & https://github.com/IgiancaUTH/HEI-MED \\

        JSIEC 
        & China & 1000
        & https://www.kaggle.com/linchundan/fundusimage1000 \\
        
        LAG 
        & China & 4854
        & https://github.com/smilell/AG-CNN \\
        
        Messidor 
        & France & 1200
        & http://www.adcis.net/en/third-party/messidor/ \\
        
        Messidor-2 
        & France & 1748
        & http://www.adcis.net/en/third-party/messidor2/ \\
        
        ODIR 
        & China & 8000
        & https://odir2019.grand-challenge.org/Download/ \\
        
        ONH Hunter 
        & UK & 99
        & http://www.aldiri.info/Image\%20Datasets/ONHSD.aspx \\
        
        Paraguay DR
        & Paraguay & 757
        & https://doi.org/10.5281/zenodo.4647952 \\
        
        RC-RGB-MA 
        & NL/China & 250
        & http://www.retinacheck.org \\

        RIGA
        & Saudi Arabia/ France & 750
        & \seqsplit{https://deepblue.lib.umich.edu/data/concern/data\_sets/3b591905z?locale=en} \\

        RIM-ONE v2
        & Spain & 455
        & http://medimrg.webs.ull.es/ \\

        RIM-ONE v3
        & Spain & 159
        & http://medimrg.webs.ull.es/ \\

        ROC 
        & NL & 100
        & http://webeye.ophth.uiowa.edu/ROC/ \\
        
        Yangxi 
        & China & 18,394
        & https://zenodo.org/record/3393265 \\
        
        \bottomrule
    \end{tabularx}
\end{table*}

\clearpage
\newpage

\begin{table*}[t]
    \centering
    \caption{Summary of Public datasets used for downstream task validation. }
    \label{tab:downstream_datasets}
    \small
    \renewcommand{\arraystretch}{1.3} 
    
    \begin{tabularx}{\textwidth}{
        >{\raggedright\arraybackslash}p{0.10\textwidth} 
        >{\raggedright\arraybackslash}p{0.12\textwidth} 
        >{\raggedright\arraybackslash}p{0.08\textwidth} 
        X
    }
        \toprule
        \textbf{Task} & \textbf{Dataset} & \textbf{Images} & \textbf{Link to Dataset} \\
        \midrule
        
        \multirow[t]{6}{=}{Vessel} 
        & CHASE\_DB1 
        & 28
        & \seqsplit{https://researchinnovation.kingston.ac.uk/en/datasets/chasedb1-retinal-vessel-reference-dataset-4/} \\
        
        & DRIVE 
        & 40
        & https://www.isi.uu.nl/Research/Databases/DRIVE/ \\
        
        & FIVES 
        & 800
        & \seqsplit{https://figshare.com/articles/figure/FIVES\_A\_Fundus\_Image\_Dataset\_for\_AI-based\_Vessel\_Segmentation/19688169} \\
        
        & HRF 
        & 45 
        & https://www5.cs.fau.de/research/data/fundus-images/ \\
        
        & IOSTAR 
        & 30 
        & http://www.retinacheck.org \\
        
        & STARE 
        & 20 
        & http://cecas.clemson.edu/\textasciitilde ahoover/stare/ \\
        \midrule
        
        \multirow[t]{7}{=}{OD / OC} 
        & Drions-DB 
        & 110 
        & http://www.ia.uned.es/\textasciitilde ejcarmona/DRIONS-DB.html \\
        
        & Drishti-GS 
        & 101
        & \seqsplit{http://cvit.iiit.ac.in/projects/mip/drishti-gs/mip-dataset2/Home.php} \\
        
        & G1020 
        & 1,020 
        & https://www.dfki.uni-kl.de/g1020 \\
        
        & GAMMA 
        & 300
        & https://gamma.grand-challenge.org/ \\
        
        & ORIGA 
        & 650 
        & \seqsplit{https://figshare.com/articles/dataset/Retinal\_Fundus\_Glaucoma\_Image\_dataset/24549217} \\
        
        & PAPILA 
        & 488 
        & https://figshare.com/articles/dataset/PAPILA/14798004 \\
        
        & REFUGE 
        & 1,200
        & https://refuge.grand-challenge.org/ \\
        \midrule
        
        \multirow[t]{4}{=}{Lesion} 
        & DDR 
        & 757
        & https://github.com/nkicsl/DDR-dataset \\
        
        & E-ophtha 
        & 463 
        & http://www.adcis.net/en/third-party/e-ophtha/ \\
        
        & FGADR 
        & 1,842
        & https://csyizhou.github.io/FGADR/ \\
        
        & IDRiD 
        & 81
        & https://idrid.grand-challenge.org/ \\
        \midrule
        
        \multirow[t]{5}{=}{General} 
        & Maples-DR 
        & 198 
        & https://github.com/LIV4D/MAPLES-DR \\
        
        & DRIVE\textsuperscript{*}
        & 40 
        & \multirow{4}{=}{Please refer to our official repository to get relabeled annotations.} \\
        
        & HRF\textsuperscript{*}
        & 45 
        & \\
        
        & GAMMA\textsuperscript{*}
        & 300 
        & \\
        
        & IDRiD\textsuperscript{*}
        & 81 
        & \\
        
        \bottomrule
        
        \multicolumn{4}{l}{\footnotesize \textsuperscript{*} These datasets were originally constructed for other tasks and have been re-annotated for the general segmentation task in this study.} \\
    \end{tabularx}
\end{table*}

\clearpage
\newpage

\begin{table*}[t]
    \centering
    \caption{Summary of datasets used for correlation analysis between fundus features and clinical diseases.}
    \label{tab:correlation_datasets}
    \small
    \renewcommand{\arraystretch}{1.3} 
    
    \begin{tabularx}{\textwidth}{
        >{\raggedright\arraybackslash}p{0.17\textwidth} 
        >{\raggedright\arraybackslash}p{0.12\textwidth} 
        >{\raggedright\arraybackslash}p{0.08\textwidth} 
        X
    }
        \toprule
        \textbf{Task} & \textbf{Dataset} & \textbf{Images} & \textbf{Link to Dataset} \\
        \midrule
        
        DR 
        & APTOS 2019
        & 3,662
        & https://www.kaggle.com/c/aptos2019-blindness-detection \\
        
        AMD 
        & ADAM 
        & 400
        & https://adam.grand-challenge.org/ \\
        
        Glaucoma 
        & REFUGE 
        & 1,200
        & https://refuge.grand-challenge.org/ \\
        
        Pathological Myopia 
        & MMAC 
        & 2,000
        & https://mmac.grand-challenge.org/ \\
        
        Multiple Diseases 
        & RFMiD 
        & 3,200
        & \seqsplit{https://ieee-dataport.org/open-access/retinal-fundus-multi-disease-image-dataset-rfmid} \\
        
        \bottomrule
    \end{tabularx}
\end{table*}

\begin{table*}[t]
    \centering
    \caption{Summary of private datasets used for task-specific expert training and adaptation.}
    \label{tab:private_datasets}
    \small
    \renewcommand{\arraystretch}{1.3} 
    
    \begin{tabularx}{\textwidth}{
        >{\raggedright\arraybackslash}p{0.12\textwidth} 
        X 
        >{\raggedright\arraybackslash}p{0.12\textwidth} 
        >{\raggedright\arraybackslash}p{0.30\textwidth} 
    }
        \toprule
        \textbf{Task} & \textbf{Target Structures} & \textbf{Images} & \textbf{Annotation Description} \\
        \midrule
        
        Vessel 
        & Artery, Vein
        & 2,509
        & Consensus of 3 Experts \\
        
        Optic Nerve 
        & Optic Disc, Optic Cup
        & 4,131
        & Single Expert \\
        
        Tessellation 
        & Tessellated Fundus
        & 4,017
        & Single Expert \\
        
        Myopia 
        & Arc Lesion, Diffuse Atrophy, Patchy Atrophy
        & 7,425
        & Single Expert \\
        
        Lesion 
        & 30 Distinct Lesion Types\textsuperscript{*}
        & 42,898
        & Single Expert \\
        
        \bottomrule
        
        \multicolumn{4}{p{\linewidth}}{\footnotesize \textsuperscript{*} \textbf{List of lesions:} Arteriovenous nicking, Artifacts, Choroidal defect, CNV hemorrhage, Cotton wool spot, Depigmentation, Diffuse atrophy, Drusen, Edema, Epiretinal membrane, Fibrous proliferation, Hard exudate, Laser spot, Macular hole, Myelinated nerve fibers, Normal foveal reflex, Patchy atrophy, Pigmentary changes, Punctate hemorrhage, Retinal detachment, Retinal neovascularization, Retinal scar, Serous detachment, Unknown abnormality, Vascular abnormality, Vascular sheathing, Venous beading, Venous tortuosity, Vitreous degeneration.} \\
        
        \multicolumn{4}{p{\linewidth}}{\footnotesize \textit{Note:} All private datasets were collected with Institutional Review Board (IRB) approval.} \\
    \end{tabularx}
\end{table*}

\begin{table*}[t]
    \centering
    \caption{Summary of cross-modal datasets used in this study. Note that PRIME-FP20 contains 15 images despite its name.}
    \label{tab:cross_modal_datasets}
    \small
    \renewcommand{\arraystretch}{1.3} 
    
    \begin{tabularx}{\textwidth}{
        >{\raggedright\arraybackslash}p{0.20\textwidth} 
        >{\raggedright\arraybackslash}p{0.18\textwidth} 
        >{\raggedright\arraybackslash}p{0.12\textwidth} 
        X
    }
        \toprule
        \textbf{Dataset} & \textbf{Modality} & \textbf{Images} & \textbf{Annotation \& Details} \\
        \midrule
        
        PRIME-FP20 
        & Ultra-wide Fundus 
        & 15
        & Contains 15 high-resolution ultra-widefield images with pixel level vessel annotations. \\
        
        Private UWF OD Seg 
        & Ultra-wide Fundus 
        & 277
        & Private dataset. Single expert annotation. Specifically annotated for OD segmentation. \\
        
        OCTA-500 
        & OCTA
        & 500
        & Used for pixel-level artery \& vein annotations. \\
        
        \bottomrule
    \end{tabularx}
\end{table*}

\clearpage
\newpage

\begin{table*}[t]
    \centering
    \caption{Quantitative comparison of different methods for vessel segmentation across six public datasets. }
    \label{tab:vessel_performance_all}
    \setlength{\tabcolsep}{3pt} 
    \renewcommand{\arraystretch}{1.25} 
    
    \resizebox{\textwidth}{!}{
        \begin{tabular}{l | ccc | ccc | ccc | ccc | ccc | ccc}
            \toprule
            \multirow{2}{*}{\textbf{Method}} 
            & \multicolumn{3}{c|}{\textbf{CHASE\_DB1}} 
            & \multicolumn{3}{c|}{\textbf{DRIVE}} 
            & \multicolumn{3}{c|}{\textbf{FIVES}} 
            & \multicolumn{3}{c|}{\textbf{HRF}} 
            & \multicolumn{3}{c|}{\textbf{IOSTAR}} 
            & \multicolumn{3}{c}{\textbf{STARE}} \\
            \cmidrule(lr){2-4} \cmidrule(lr){5-7} \cmidrule(lr){8-10} \cmidrule(lr){11-13} \cmidrule(lr){14-16} \cmidrule(lr){17-19}
             & DSC & JAC & clDice & DSC & JAC & clDice & DSC & JAC & clDice & DSC & JAC & clDice & DSC & JAC & clDice & DSC & JAC & clDice \\
            \midrule
            
            MedSAM 
            & 0.179 & 0.098 & 0.811 & 0.194 & 0.108 & 0.870 & 0.157 & 0.086 & 0.949 & 0.159 & 0.086 & 0.750 & 0.172 & 0.094 & 0.983 & 0.177 & 0.098 & 0.845 \\
            
            MedSAM2 
            & 0.744 & 0.594 & 0.228 & 0.666 & 0.501 & \textbf{0.438} & 0.778 & 0.641 & 0.312 & 0.547 & 0.379 & \textbf{0.639} & 0.717 & 0.560 & 0.231 & 0.735 & 0.581 & 0.296 \\
            
            Med-SAM-Adapter 
            & 0.715 & 0.557 & 0.217 & 0.712 & 0.553 & 0.234 & 0.718 & 0.563 & 0.202 & 0.544 & 0.375 & 0.431 & 0.688 & 0.527 & 0.159 & 0.658 & 0.491 & 0.233 \\
            
            SAM2 
            & 0.772 & 0.629 & 0.221 & 0.755 & 0.607 & 0.291 & 0.781 & 0.656 & 0.249 & 0.725 & 0.570 & 0.371 & 0.760 & 0.614 & 0.160 & 0.784 & 0.646 & 0.193 \\
            
            SAM3 
            & 0.784 & 0.638 & 0.206 & 0.758 & 0.614 & 0.256 & 0.816 & 0.704 & 0.205 & 0.717 & 0.559 & 0.361 & 0.763 & 0.619 & 0.160 & 0.781 & 0.641 & 0.185 \\
            
            \midrule
            
            SAM2-UNet 
            & 0.793 & 0.657 & 0.199 & 0.766 & 0.621 & 0.231 & 0.864 & 0.766 & 0.117 & 0.699 & 0.538 & 0.351 & 0.766 & 0.621 & 0.144 & 0.778 & 0.637 & 0.173 \\
            
            RetFound 
            & 0.665 & 0.498 & 0.261 & 0.522 & 0.354 & 0.372 & 0.787 & 0.655 & 0.253 & 0.576 & 0.405 & 0.549 & 0.637 & 0.469 & 0.299 & 0.548 & 0.378 & 0.340 \\
            
            \midrule
            
            RetSAM - Linear 
            & 0.804 & \textbf{0.674} & 0.259 & 0.793 & 0.658 & 0.345 & 0.854 & 0.753 & 0.329 & 0.779 & 0.640 & 0.418 & 0.772 & 0.629 & 0.341 & 0.818 & 0.693 & 0.311 \\
            
            RetSAM - Finetune 
            & \textbf{0.815} & 0.655 & \textbf{0.266} & \textbf{0.813} & \textbf{0.686} & 0.368 & \textbf{0.901} & \textbf{0.827} & \textbf{0.341} & \textbf{0.799} & \textbf{0.667} & 0.390 & \textbf{0.781} & \textbf{0.642} & \textbf{0.351} & \textbf{0.824} & \textbf{0.701} & \textbf{0.362} \\
            
            \bottomrule
        \end{tabular}
    }
\end{table*}

\begin{table*}[t]
    \centering
    \caption{Quantitative comparison of different methods for OD segmentation across six public datasets.}
    \label{tab:od_performance}
    \setlength{\tabcolsep}{1.5pt} 
    \renewcommand{\arraystretch}{1.25}
    
    \resizebox{\textwidth}{!}{
        \begin{tabular}{l | ccc | ccc | ccc | ccc | ccc | ccc | ccc}
            \toprule
            \multirow{2}{*}{\textbf{Method}} 
            & \multicolumn{3}{c|}{\textbf{Drions-DB}} 
            & \multicolumn{3}{c|}{\textbf{Drishti-GS}} 
            & \multicolumn{3}{c|}{\textbf{G1020}} 
            & \multicolumn{3}{c|}{\textbf{Gamma}} 
            & \multicolumn{3}{c|}{\textbf{Origa}} 
            & \multicolumn{3}{c|}{\textbf{Papila}} 
            & \multicolumn{3}{c}{\textbf{Refuge}} \\
            \cmidrule(lr){2-4} \cmidrule(lr){5-7} \cmidrule(lr){8-10} \cmidrule(lr){11-13} \cmidrule(lr){14-16} \cmidrule(lr){17-19} \cmidrule(lr){20-22}
             & DSC & JAC & HD95 & DSC & JAC & HD95 & DSC & JAC & HD95 & DSC & JAC & HD95 & DSC & JAC & HD95 & DSC & JAC & HD95 & DSC & JAC & HD95 \\
            \midrule
            
            MedSAM 
            & 0.964 & 0.932 & 4.01 & 0.954 & 0.912 & 17.44 & 0.938 & 0.888 & 28.35 & 0.947 & 0.902 & 14.53 & 0.955 & 0.916 & 16.64 & \textbf{0.963} & \textbf{0.929} & 19.07 & \textbf{0.980} & \textbf{0.961} & 5.21 \\
            
            MedSAM2 
            & \textbf{0.974} & \textbf{0.949} & \textbf{3.09} & \textbf{0.971} & \textbf{0.945} & 13.71 & 0.936 & 0.857 & 20.82 & 0.953 & 0.910 & 14.56 & \textbf{0.971} & \textbf{0.945} & 10.50 & \textbf{0.963} & 0.928 & 19.55 & 0.955 & 0.914 & 10.98 \\
            
            Med-SAM-Adapter 
            & 0.960 & 0.923 & 4.89 & 0.917 & 0.852 & 28.07 & 0.820 & 0.738 & 65.39 & 0.815 & 0.721 & 69.44 & 0.800 & 0.702 & 45.50 & 0.853 & 0.786 & 51.45 & 0.806 & 0.706 & 38.38 \\
            
            SAM2 
            & 0.965 & 0.933 & 3.66 & 0.965 & 0.933 & 15.49 & \textbf{0.951} & \textbf{0.908} & 23.14 & 0.949 & 0.905 & 14.56 & 0.965 & 0.933 & 13.50 & 0.955 & 0.914 & 23.57 & 0.958 & 0.919 & 10.89 \\
            
            SAM3 
            & 0.968 & 0.939 & 3.55 & 0.968 & 0.936 & 15.15 & 0.950 & 0.904 & 24.30 & 0.949 & 0.904 & 14.44 & 0.967 & 0.936 & 12.70 & 0.953 & 0.910 & 24.40 & 0.955 & 0.912 & 22.96 \\
            
            \midrule
            
            SAM2-UNet 
            & 0.972 & 0.946 & 3.28 & \textbf{0.971} & 0.944 & 14.35 & 0.945 & 0.898 & 26.87 & 0.948 & 0.902 & 13.91 & 0.970 & 0.942 & 10.82 & 0.950 & 0.906 & 26.26 & 0.951 & 0.907 & 46.70 \\
            
            RetFound 
            & 0.958 & 0.921 & 5.26 & 0.961 & 0.926 & 32.32 & 0.924 & 0.862 & 33.62 & 0.938 & 0.884 & 18.36 & 0.960 & 0.924 & 15.58 & 0.937 & 0.883 & 33.24 & 0.944 & 0.896 & 18.74 \\
            
            \midrule
            
            RetSAM - Linear 
            & 0.964 & 0.936 & 5.82 & 0.947 & 0.900 & 6.28 & 0.938 & 0.888 & \textbf{6.96} & 0.953 & 0.910 & 5.17 & 0.949 & 0.904 & \textbf{5.43} & 0.937 & 0.885 & \textbf{9.35} & 0.941 & 0.890 & 5.41 \\
            
            RetSAM - Finetune 
            & 0.967 & 0.939 & 5.76 & \textbf{0.971} & 0.939 & \textbf{3.59} & 0.940 & 0.883 & 8.42 & \textbf{0.974} & \textbf{0.949} & \textbf{3.83} & 0.960 & 0.919 & 6.71 & 0.958 & 0.911 & 10.66 & 0.967 & 0.938 & \textbf{3.27} \\
            
            \bottomrule
        \end{tabular}
    }
\end{table*}

\begin{table*}[t]
    \centering
    \caption{Quantitative comparison of different methods for OC segmentation across six public datasets.}
    \label{tab:oc_performance}
    \setlength{\tabcolsep}{3pt} 
    \renewcommand{\arraystretch}{1.25}
    
    \resizebox{\textwidth}{!}{
        \begin{tabular}{l | ccc | ccc | ccc | ccc | ccc | ccc}
            \toprule
            \multirow{2}{*}{\textbf{Method}} 
            & \multicolumn{3}{c|}{\textbf{Drishti-GS}} 
            & \multicolumn{3}{c|}{\textbf{G1020}} 
            & \multicolumn{3}{c|}{\textbf{Gamma}} 
            & \multicolumn{3}{c|}{\textbf{Origa}} 
            & \multicolumn{3}{c|}{\textbf{Papila}} 
            & \multicolumn{3}{c}{\textbf{Refuge}} \\
            \cmidrule(lr){2-4} \cmidrule(lr){5-7} \cmidrule(lr){8-10} \cmidrule(lr){11-13} \cmidrule(lr){14-16} \cmidrule(lr){17-19}
             & DSC & JAC & HD95 & DSC & JAC & HD95 & DSC & JAC & HD95 & DSC & JAC & HD95 & DSC & JAC & HD95 & DSC & JAC & HD95 \\
            \midrule
            
            MedSAM 
            & 0.846 & 0.743 & 35.33 & 0.805 & 0.685 & 34.88 & 0.841 & 0.731 & 22.43 & 0.835 & 0.727 & 31.22 & 0.841 & 0.737 & 24.85 & \textbf{0.963} & \textbf{0.930} & 4.68 \\
            
            MedSAM2 
            & 0.896 & 0.815 & 27.75 & 0.906 & 0.831 & 17.70 & 0.868 & 0.771 & 18.33 & \textbf{0.925} & \textbf{0.862} & 15.89 & 0.848 & 0.745 & 20.73 & 0.862 & 0.761 & 13.90 \\
            
            Med-SAM-Adapter 
            & 0.836 & 0.739 & 40.81 & 0.760 & 0.631 & 54.74 & 0.613 & 0.461 & 96.01 & 0.844 & 0.741 & 34.08 & 0.661 & 0.535 & 73.42 & 0.751 & 0.612 & 30.47 \\
            
            SAM2 
            & 0.906 & 0.829 & 27.18 & 0.881 & 0.789 & 24.87 & 0.869 & 0.770 & 21.25 & 0.894 & 0.810 & 24.22 & 0.858 & 0.757 & 23.96 & 0.869 & 0.771 & 15.56 \\
            
            SAM3 
            & 0.905 & 0.827 & 27.68 & 0.860 & 0.764 & 27.49 & 0.865 & 0.765 & 20.73 & 0.891 & 0.807 & 23.70 & 0.835 & 0.725 & 26.58 & 0.858 & 0.749 & 17.86 \\
            
            \midrule
            
            SAM2-UNet 
            & 0.901 & 0.825 & 28.90 & 0.810 & 0.703 & 37.20 & 0.854 & 0.754 & 19.53 & 0.888 & 0.804 & 21.84 & 0.783 & 0.673 & 31.50 & 0.828 & 0.714 & 20.06 \\
            
            RetFound 
            & 0.880 & 0.801 & 31.34 & 0.764 & 0.644 & 64.66 & 0.778 & 0.668 & 22.69 & 0.862 & 0.766 & 27.71 & 0.617 & 0.507 & 41.29 & 0.856 & 0.755 & 16.72 \\
            
            \midrule
            
            RetSAM - Linear 
            & 0.782 & 0.658 & 17.51 & 0.761 & 0.639 & 10.20 & 0.742 & 0.617 & 10.15 & 0.784 & 0.664 & 10.91 & 0.794 & 0.680 & 11.74 & 0.793 & 0.672 & 7.72 \\
            
            RetSAM - Finetune 
            & \textbf{0.942} & \textbf{0.908} & \textbf{4.25} & \textbf{0.911} & \textbf{0.842} & \textbf{9.97} & \textbf{0.914} & \textbf{0.844} & \textbf{5.21} & 0.918 & 0.846 & \textbf{8.64} & \textbf{0.856} & \textbf{0.746} & \textbf{9.88} & 0.919 & 0.853 & \textbf{3.79} \\
            
            \bottomrule
        \end{tabular}
    }
\end{table*}

\clearpage
\newpage

\begin{table*}[t]
\centering
\caption{Quantitative comparison of different methods for lesion segmentation across four datasets. Results are reported in DSC. ``-'' denotes failed or invalid segmentation.}
\label{tab:lesion_dsc_transposed}
\setlength{\tabcolsep}{2.5pt} 
\tiny 
\renewcommand{\arraystretch}{1.2}
\resizebox{\textwidth}{!}{
\begin{tabular}{l | cccc | cc | cccc | cccc}
\toprule
\multirow{2}{*}{\textbf{Method}} & \multicolumn{4}{c|}{\textbf{DDR}} & \multicolumn{2}{c|}{\textbf{E-ophtha}} & \multicolumn{4}{c|}{\textbf{FGADR}} & \multicolumn{4}{c}{\textbf{IDRiD}} \\
\cmidrule(lr){2-5} \cmidrule(lr){6-7} \cmidrule(lr){8-11} \cmidrule(lr){12-15}
 & EX & HE & SE & MA & EX & MA & EX & HE & SE & MA & EX & HE & SE & MA \\
\midrule
MedSAM & 0.081 & 0.066 & 0.288 & 0.021 & 0.193 & 0.001 & 0.166 & 0.125 & 0.289 & 0.081 & 0.054 & 0.071 & 0.315 & 0.004 \\
MedSAM2 & 0.457 & 0.443 & \textbf{0.532} & 0.222 & 0.566 & 0.239 & \textbf{0.529} & \textbf{0.571} & \textbf{0.633} & 0.308 & 0.425 & 0.408 & \textbf{0.660} & 0.079 \\
Med-SAM-Adapter & 0.381 & 0.301 & 0.310 & 0.118 & 0.447 & 0.042 & 0.404 & 0.402 & 0.397 & 0.127 & 0.319 & 0.104 & 0.242 & 0.032 \\
SAM2 & 0.528 & 0.404 & 0.387 & 0.287 & 0.528 & 0.396 & 0.491 & 0.527 & 0.545 & 0.262 & 0.420 & 0.233 & 0.453 & 0.069 \\
SAM3 & 0.517 & 0.413 & 0.505 & \textbf{0.335} & 0.504 & 0.356 & 0.502 & 0.524 & 0.590 & 0.250 & 0.465 & 0.304 & 0.420 & 0.233 \\
SAM2-UNet & 0.443 & 0.397 & - & - & 0.518 & 0.137 & 0.417 & 0.465 & - & - & 0.558 & 0.504 & 0.298 & \textbf{0.616} \\
RetFound & 0.234 & 0.231 & 0.309 & - & 0.296 & 0.001 & 0.303 & 0.346 & 0.355 & 0.032 & 0.303 & 0.358 & 0.308 & - \\
\midrule
RetSAM - Linear & 0.431 & 0.402 & 0.445 & - & 0.527 & - & 0.316 & 0.414 & 0.436 & - & 0.538 & 0.482 & 0.527 & - \\
RetSAM - Finetune & \textbf{0.578} & \textbf{0.465} & 0.518 & 0.314 & \textbf{0.567} & \textbf{0.404} & 0.508 & 0.563 & 0.601 & \textbf{0.344} & \textbf{0.606} & \textbf{0.515} & 0.595 & 0.395 \\
\bottomrule
\end{tabular}
}
\end{table*}

\begin{table*}[t]
\centering
\caption{Quantitative comparison of different methods for lesion segmentation across four datasets. Results are reported in Precision. ``-'' denotes failed or invalid segmentation.}
\label{tab:lesion_precision_transposed}
\setlength{\tabcolsep}{2.5pt}
\tiny 
\renewcommand{\arraystretch}{1.2}
\resizebox{\textwidth}{!}{
\begin{tabular}{l | cccc | cc | cccc | cccc}
\toprule
\multirow{2}{*}{\textbf{Method}} & \multicolumn{4}{c|}{\textbf{DDR}} & \multicolumn{2}{c|}{\textbf{E-ophtha}} & \multicolumn{4}{c|}{\textbf{FGADR}} & \multicolumn{4}{c}{\textbf{IDRiD}} \\
\cmidrule(lr){2-5} \cmidrule(lr){6-7} \cmidrule(lr){8-11} \cmidrule(lr){12-15}
 & EX & HE & SE & MA & EX & MA & EX & HE & SE & MA & EX & HE & SE & MA \\
\midrule
MedSAM & 0.056 & 0.046 & 0.256 & 0.015 & 0.141 & 0.001 & 0.150 & 0.115 & 0.311 & 0.067 & 0.030 & 0.055 & 0.316 & 0.002 \\
MedSAM2 & 0.481 & 0.549 & 0.570 & 0.203 & 0.523 & 0.244 & 0.511 & 0.529 & \textbf{0.621} & 0.322 & 0.401 & 0.393 & \textbf{0.772} & 0.051 \\
Med-SAM-Adapter & 0.417 & 0.545 & 0.377 & 0.215 & 0.375 & 0.044 & 0.456 & 0.530 & 0.490 & 0.240 & 0.541 & 0.274 & 0.442 & 0.088 \\
SAM2 & 0.573 & 0.501 & 0.451 & 0.251 & 0.615 & 0.420 & 0.516 & 0.571 & 0.537 & 0.207 & 0.512 & 0.263 & 0.554 & 0.179 \\
SAM3 & 0.543 & 0.495 & 0.506 & \textbf{0.303} & 0.589 & 0.378 & 0.516 & 0.560 & 0.575 & 0.237 & 0.498 & 0.328 & 0.468 & 0.295 \\
SAM2-UNet & 0.410 & 0.476 & - & - & 0.459 & 0.111 & 0.400 & 0.453 & - & - & 0.475 & 0.506 & 0.264 & \textbf{0.653} \\
RetFound & 0.275 & 0.367 & 0.354 & - & 0.336 & 0.001 & 0.288 & 0.371 & 0.430 & 0.056 & 0.221 & 0.470 & 0.538 & - \\
\midrule
RetSAM - Linear & 0.546 & 0.465 & 0.106 & - & 0.634 & - & 0.481 & 0.538 & 0.196 & - & 0.615 & \textbf{0.513} & 0.405 & - \\
RetSAM - Finetune & \textbf{0.592} & \textbf{0.606} & \textbf{0.589} & 0.199 & \textbf{0.702} & \textbf{0.441} & \textbf{0.577} & \textbf{0.569} & 0.591 & \textbf{0.228} & \textbf{0.630} & 0.505 & 0.562 & 0.327 \\
\bottomrule
\end{tabular}
}
\end{table*}

\clearpage
\newpage

\begin{table*}[t]
    \centering
    \caption{Quantitative comparison of different methods for general segmentation ability across four relabeled public datasets. Results are reported in DSC. Datasets/lesions in columns. ``-'' denotes failed or invalid segmentation.}
    \label{tab:general_performance_part1}
    \setlength{\tabcolsep}{1.5pt} 
    \renewcommand{\arraystretch}{1.25}
    
    \resizebox{\textwidth}{!}{
        \begin{tabular}{l | ccccc | ccccc | ccccc | ccccc}
            \toprule
            \multirow{2}{*}{\textbf{Method}} 
            & \multicolumn{5}{c|}{\textbf{DRIVE}} 
            & \multicolumn{5}{c|}{\textbf{GAMMA}} 
            & \multicolumn{5}{c|}{\textbf{HRF}} 
            & \multicolumn{5}{c}{\textbf{IDRiD}} \\
            \cmidrule(lr){2-6} \cmidrule(lr){7-11} \cmidrule(lr){12-16} \cmidrule(lr){17-21}
             & Vessel & OD & OC & EX & HE & Vessel & OD & OC & EX & HE & Vessel & OD & OC & EX & HE & Vessel & OD & OC & EX & HE \\
            \midrule
            
            MedSAM 
            & 0.129 & 0.828 & 0.505 & 0.002 & 0.001 & 0.095 & 0.924 & 0.763 & 0.001 & - & 0.144 & 0.926 & 0.735 & 0.003 & 0.003 & 0.118 & 0.919 & 0.809 & 0.022 & 0.016 \\
            
            MedSAM2 
            & 0.462 & 0.845 & 0.515 & - & 0.001 & 0.439 & 0.943 & 0.778 & - & - & 0.485 & 0.946 & 0.749 & 0.257 & 0.003 & 0.469 & 0.938 & 0.826 & 0.417 & 0.015 \\
            
            Med-SAM-Adapter 
            & 0.047 & 0.762 & 0.367 & - & - & 0.037 & 0.905 & 0.698 & - & - & 0.016 & 0.927 & 0.668 & - & 0.072 & 0.073 & 0.718 & 0.526 & 0.013 & 0.028 \\
            
            SAM2
            & 0.644 & 0.914 & 0.726 & 0.043 & 0.100 & 0.710 & 0.961 & 0.737 & 0.220 & 0.394 & 0.583 & 0.942 & 0.782 & 0.217 & 0.006 & 0.537 & 0.958 & 0.756 & 0.458 & 0.289 \\
            
            SAM3 
            & 0.655 & 0.925 & 0.748 & 0.085 & 0.140 & 0.720 & 0.966 & 0.748 & 0.225 & 0.402 & 0.592 & 0.949 & 0.792 & 0.220 & 0.007 & 0.545 & 0.962 & 0.763 & 0.465 & 0.295 \\
            
            \midrule
            
            SAM2-UNet 
            & 0.659 & 0.934 & 0.789 & - & - & 0.600 & 0.960 & 0.893 & - & - & 0.633 & 0.968 & 0.805 & 0.292 & 0.266 & 0.688 & 0.968 & 0.843 & 0.568 & 0.511 \\
            
            RetFound 
            & 0.579 & 0.895 & 0.540 & - & - & 0.580 & 0.949 & 0.784 & - & - & 0.631 & 0.955 & 0.778 & - & - & 0.623 & 0.964 & 0.777 & 0.532 & 0.430 \\
            
            \midrule
            
            RetSAM-Linear 
            & 0.793 & 0.957 & 0.904 & 0.358 & 0.175 & 0.725 & 0.954 & 0.617 & 0.482 & 0.450 & 0.779 & 0.969 & 0.915 & 0.337 & 0.359 & 0.764 & 0.974 & \textbf{0.932} & 0.586 & \textbf{0.589} \\
            
            RetSAM-Finetune 
            & \textbf{0.828} & \textbf{0.972} & \textbf{0.911} & \textbf{0.469} & \textbf{0.238} & \textbf{0.886} & \textbf{0.984} & \textbf{0.955} & \textbf{0.732} & \textbf{0.682} & \textbf{0.804} & \textbf{0.986} & \textbf{0.945} & \textbf{0.425} & \textbf{0.401} & \textbf{0.875} & \textbf{0.988} & 0.930 & \textbf{0.639} & 0.550 \\
            
            \bottomrule
        \end{tabular}
    }
\end{table*}

\begin{table*}[t]
    \centering
    \caption{Quantitative comparison of different methods for general segmentation ability on Maples-DR. Results are reported in DSC. Datasets/lesions in columns. ``-'' denotes failed or invalid segmentation.}
    \label{tab:general_performance_maples}
    \setlength{\tabcolsep}{2.5pt} 
    \renewcommand{\arraystretch}{1.25}
    
    \resizebox{0.5\textwidth}{!}{ 
        \begin{tabular}{l | ccccccc}
            \toprule
            \multirow{2}{*}{\textbf{Method}} 
            & \multicolumn{7}{c}{\textbf{Maples-DR}} \\
            \cmidrule(lr){2-8}
             & Vessel & OD & OC & EX & HE & SE & Drusen \\
            \midrule
            
            MedSAM 
            & - & 0.899 & 0.655 & 0.026 & 0.001 & - & - \\
            
            MedSAM2 
            & 0.001 & 0.918 & 0.668 & 0.151 & 0.002 & - & - \\
            
            Med-SAM-Adapter
            & 0.194 & 0.921 & 0.681 & 0.013 & 0.036 & 0.045 & - \\
            
            SAM2
            & 0.693 & 0.946 & 0.704 & 0.322 & 0.285 & 0.464 & 0.155 \\
            
            SAM3 
            & 0.703 & 0.905 & 0.710 & 0.330 & 0.292 & \textbf{0.472} & 0.158 \\
            
            \midrule
            
            SAM2-UNet 
            & 0.643 & 0.170 & 0.132 & 0.295 & 0.027 & - & - \\
            
            RetFound 
            & 0.544 & 0.928 & 0.662 & 0.148 & 0.058 & - & 0.057 \\
            
            \midrule
            
            RetSAM-Linear 
            & 0.762 & 0.912 & 0.617 & 0.196 & 0.246 & 0.202 & 0.199 \\
            
            RetSAM-Finetune 
            & \textbf{0.853} & \textbf{0.957} & \textbf{0.828} & \textbf{0.398} & \textbf{0.374} & \textbf{0.472} & \textbf{0.401} \\
            
            \bottomrule
        \end{tabular}
    }
\end{table*}

\clearpage
\newpage

\begin{table*}[t]
    \centering
    \caption{Quantitative comparisons of RetSAM against single-task expert models on the private datasets used for pseudo-label generation. Results are reported in DSC. }
    \label{tab:ablation_study}
    \setlength{\tabcolsep}{3pt} 
    \renewcommand{\arraystretch}{1.25}
    
    \resizebox{\textwidth}{!}{
        \begin{tabular}{l | c c c c c c c c c}
            \toprule
            \textbf{Method} & \textbf{Artery} & \textbf{Vein} & \textbf{OD} & \textbf{OC} & \textbf{Tessellation} & \textbf{Arc Lesion} & \textbf{Diffuse Atrophy} & \textbf{Patchy Atrophy} & \textbf{Lesion Avg.} \\
            \midrule
            
            Baseline 
            & 0.627 & 0.699 & 0.939 & 0.786 & 0.528 & 0.705 & 0.619 & 0.497 & 0.322 \\
            
            RetSAM - Linear 
            & 0.662 & 0.727 & 0.944 & 0.845 & 0.559 & 0.757 & 0.637 & 0.528 & 0.336 \\
            
            RetSAM - Finetune 
            & \textbf{0.687} & \textbf{0.740} & \textbf{0.956} & \textbf{0.886} & \textbf{0.597} & \textbf{0.783} & \textbf{0.695} & \textbf{0.550} & \textbf{0.396} \\
            
            \bottomrule
        \end{tabular}
    }
\end{table*}

\begin{table*}[t]
    \centering
    \caption{Cross-modal generalization assessment across distinct imaging modalities. Results are reported in DSC. Due to the substantial domain shift, RetSAM was unable to produce valid segmentations on OCTA500 under the linear inference setting. }
    \label{tab:cross_modal_performance}
    \small 
    \setlength{\tabcolsep}{6pt} 
    \renewcommand{\arraystretch}{1.25}
    
    \begin{tabular}{l | c | c | c c}
        \toprule
        \multirow{2}{*}{\textbf{Method}} 
        & \textbf{Prime-FP20} 
        & \textbf{Private UWF} 
        & \multicolumn{2}{c}{\textbf{OCTA-500}} \\
        \cmidrule(lr){2-2} \cmidrule(lr){3-3} \cmidrule(lr){4-5}
         & Vessel & OD & Artery & Vein \\
        \midrule
        
        Baseline
        & 0.230 & 0.734 & 0.617 & 0.625 \\
        
        SAM2-UNet 
        & 0.594 & 0.752 & 0.631 & 0.646 \\
        
        RetSAM - Linear 
        & 0.684 & 0.776 & - & - \\
        
        RetSAM - Finetune 
        & \textbf{0.737} & \textbf{0.833} & \textbf{0.666} & \textbf{0.676} \\
        
        \bottomrule
    \end{tabular}
\end{table*}

\clearpage
\newpage

\begin{table*}[t]
\centering
\caption{Summary of quantitative oculomics features and biomarkers extracted by the proposed framework.}
\label{tab:feature_definitions}
\renewcommand{\arraystretch}{1.35} 
\begin{tabularx}{\textwidth}{l l X}
\toprule
Category & Metric / Feature & Description \\
\midrule

\multirow[t]{4}{*}{Retinal Vessels} 
 & A/V Ratio & Mean diameter ratio between arteries and veins. \\
 & CRAE / CRVE & Central Retinal Artery/Vein Equivalents (vascular caliber). \\
 & Fractal Dimension & Branching complexity index for arteries ($FD_a$) and veins ($FD_v$). \\
 & Tortuosity & Geometrical curvature measure quantifying vessel twist. \\
\midrule

\multirow[t]{5}{3.5cm}{Optic Disc, Cup \\ \& Macula} 
 & Cup-to-Disc Ratio & Horizontal (CDR) and Vertical (vCDR) diameter ratios. \\
 & ISNT Parameters & Neuroretinal rim widths in Inferior, Superior, Nasal, Temporal sectors. \\
 & Orientation & Angle of the major axis of the disc/cup relative to the horizontal. \\
 & Foveal Localization & Coordinates of the macula fovea center (pixels). \\
 & Morphometry & Pixel area measurements for optic disc and cup. \\
\midrule

\multirow[t]{3}{*}{Tessellation} 
 & Coverage Ratio & Ratio of tessellated fundus area to the total analyzable area. \\
 & Shape Descriptors & Mean circularity and aspect ratio describing texture shape. \\
 & Centroid Dispersion & Spatial dispersion metric of tessellation component centroids. \\
\midrule

\multirow[t]{2}{3.5cm}{Pathological \\ Myopia} 
 & Atrophy Metrics & Count, area, and coverage ratio for Diffuse, Patchy, and PPA. \\
 & Global Coverage & Aggregated coverage ratio of all myopia-related structural changes. \\
\midrule

\multirow[t]{5}{*}{Lesion} 
 & Lesion Load & Total count, pixel area, and coverage ratio per lesion category. \\
 & Size Distribution & Counts of lesions stratified by size (Small, Medium, Large). \\
 & Shape Morphology & Geometric metrics including Circularity and Aspect Ratio. \\
 & Spatial Localization & Lesion counts per macula-centered quadrant. \\
 & Severity Grading & Automated severity grading based on lesion coverage ratio. \\

\bottomrule
\end{tabularx}
\end{table*}

\clearpage
\newpage

\begin{table}[h]
\centering
\caption{Implementation details and hyperparameter settings for the proposed RetSAM.}
\label{tab:implementation_details}
\renewcommand{\arraystretch}{1.2} 
\begin{tabular}{lll}
\toprule
\textbf{Category} & \textbf{Parameter} & \textbf{Value} \\
\midrule
\multirow[t]{8}{*}{Architecture} 
 & Backbone & Swin-Base \\
 & Pre-training Weights & ImageNet-22K \\
 & Input Resolution & $640 \times 640$ \\
 & Patch Size & 4 \\
 & Window Size & 10 \\
 & Embedding Dim. ($C$) & 128 \\
 & Depths & $[2, 2, 18, 2]$ \\
 & Attention Heads & $[4, 8, 16, 32]$ \\
\midrule
\multirow[t]{7}{*}{Training} 
 & Optimizer & AdamW \\
 & Initial Learning Rate & $1 \times 10^{-4}$ \\
 & Weight Decay & $0.01$ \\
 & Optimizer Betas ($\beta_1, \beta_2$) & $(0.9, 0.999)$ \\
 & Batch Size & 12 \\
 & Total Epochs & 100 \\
\bottomrule
\end{tabular}
\end{table}

\clearpage
\newpage

\section*{Methods}
\label{sec:Methods}

\subsection*{Segmentation Task Formulation and Taxonomy}

Existing retinal datasets are fragmented and typically focus on isolated objectives, such as vessel segmentation or optic disc/cup segmentation. To address this limitation, RetSAM integrates these targets into a unified task taxonomy to support standardized quantitative phenotyping for ophthalmology and oculomics. By harmonizing annotations from diverse sources, we define a comprehensive set of segmentation targets that enables a single model to segment anatomical, pathological, and phenotypic features. We organize these targets into three semantic groups based on their clinical and biological characteristics.

\paragraph{Anatomical Structures} This category includes core structures that are expected in all valid fundus images, independent of disease status. Although their appearance varies across individuals, these structures follow consistent anatomical organization. Examples include retinal vessels, further distinguished into arteries and veins, as well as the optic disc and optic cup. Accurate segmentation of these structures provides the geometric reference for downstream quantification, such as the cup-to-disc ratio in glaucoma.

\paragraph{Lesions} Lesions represent focal pathological changes that are absent in healthy eyes. We define this category to cover a broad spectrum of abnormalities. In addition to common signs of major retinopathies, including microaneurysms, hemorrhages, hard exudates, cotton-wool spots, and drusen, this category also includes targets such as fibrous proliferation and other non-specific anomalies. These lesions are morphologically diverse, ranging from small punctate spots to large confluent regions, with heterogeneous intensity and texture on fundus imaging. Precise lesion segmentation is clinically important for characterizing diseases such as diabetic retinopathy and age-related macular degeneration, and for capturing signs associated with hypertensive retinopathy, retinal vein occlusion, and broader systemic conditions.

\paragraph{Fundus Phenotypes} Fundus phenotypes refer to visual patterns that are not acute focal lesions but are associated with anatomical states or disease risk. Phenotypes such as tessellated fundus and peripapillary atrophy represent chronic or structural characteristics of the retinal background. They often exhibit diffuse boundaries and distinctive textural patterns. Quantifying these patterns supports risk profiling and disease characterization, such as assessing myopia severity or capturing changes relevant to glaucoma.

\paragraph{}For the complete list of segmentation tasks defined in RetSAM, see Table~\ref{tab:segmentation_tasks}. Because source datasets often provide partial labels, many training samples are only annotated for a subset of targets. We address this setting with a training strategy tailored to partial supervision, described in the Training Strategy section.

\subsection*{Datasets}

\subsubsection*{Hybrid Training Corpus}

To ensure both clinical precision and robustness, we construct a hybrid training corpus comprising two distinct yet complementary sources.

\paragraph{High-Fidelity In-house Cohorts} To secure reliable supervision, we curate a series of proprietary clinical datasets derived from Airdoc's online fundus screening services from 2018 to 2024. The data acquisition, cleansing, and expert annotation workflow is performed by the Airdoc Medical Department. These cohorts serve as the ground-truth backbone, comprising over 50{,}000 high-resolution images. Following IRB approval and automated de-identification, all in-house cohorts are partitioned using a consistent strategy: 70\% for training, 10\% for validation, and 20\% for internal testing. A summary of the datasets is provided in Table~\ref{tab:private_datasets}. We construct four specialized subsets.

\noindent \textbf{(1) Artery and Vein Segmentation.} Distinguishing arteries from veins is essential for blood flow analysis. We collected 2{,}509 images specifically for this task. To ensure reliable evaluation, the training set was labeled by a single senior expert, while the test set was independently annotated by three ophthalmologists. Disagreements in the test set were resolved through discussion to establish a consensus ground truth.

\noindent \textbf{(2) Optic Disc and Cup Segmentation.} To support the diagnosis of glaucoma and other optic nerve diseases, we collected 4{,}131 images targeting the OD and OC. Senior ophthalmologists annotated these structures following standard clinical protocols, ensuring that the boundaries of the neuroretinal rim are strictly defined.

\noindent \textbf{(3) Pathological Lesion Segmentation.} To support precise disease quantification, we constructed a large dataset of 42{,}898 images with pixel-level annotations for 29 distinct lesion types. Unlike public datasets that typically focus on single diseases, this cohort covers a wide range of lesions, from common DR and AMD signs to rare abnormalities. Since rare lesions often have too few samples for effective training, we group them into a unified class termed Possible Lesion. This class serves as an alert signal, highlighting suspicious regions that contain pathological changes but lack sufficient evidence for a specific classification, so that these areas can be flagged for clinical review.

\noindent \textbf{(4) Myopic and Structural Phenotyping.} Given the growing prevalence of myopia, we curate two specialized datasets to capture structural complications: one for fundus tessellation and another for myopic features. These datasets target distinct stages of myopic maculopathy, encompassing tessellation, PPA, diffuse atrophy, and patchy atrophy. By focusing on these morphological markers, the datasets capture subtle texture changes and geographic atrophy zones associated with choroidal thinning and pathological myopia progression.

\paragraph{Large-Scale Open-Source Integration} We aggregate 24 publicly available datasets \cite{diaz2019cnns,li2019diagnostic,liu2022deepdrid,kauppi2007diaretdb1,cuadros2009eyepacs,hernandez2017fire,giancardo2012exudate,cen2021automatic,li2019attention,decenciere2013teleophta,decenciere2014feedback,li2020benchmark,almazroa2018retinal,fumero2011rim,niemeijer2009retinopathy,islam2021deep,pires2014advancing,lowell2004measurement,benitez2021dataset,liu2019self,dashtbozorg2018retinal}, comprising approximately 160{,}000 color fundus images, with details listed in Table~\ref{tab:public_datasets}. This large collection represents a diverse real-world clinical environment. It spans populations across Asia, Europe, and the Americas and covers a broad disease spectrum ranging from common conditions such as DR and AMD to rare pathologies. It also introduces technical variability through differences in device manufacturers such as Zeiss, Canon, and Topcon, as well as resolutions and illumination conditions. This corpus serves to populate our unified task taxonomy. Because we treat these public datasets as unlabeled data due to the lack of unified supervision, we use models trained on our high-fidelity in-house data to distill knowledge into this collection. Specifically, we generate pseudo-labels for these public images and map the visual data into our comprehensive label space. Given its role in representation learning, we adopt a uniform splitting strategy, allocating 90\% of the images for training and 10\% for validation. We do not establish a held-out test set, as we do not report quantitative metrics on this compilation.

\subsubsection*{Multi-Dimensional Evaluation Benchmarks}

To comprehensively assess RetSAM's capabilities, we establish a three-tier evaluation framework ranging from standardized single-task benchmarking to unified multi-task assessment and cross-modal adaptation testing. Full details are summarized in Tables~\ref{tab:downstream_datasets} and~\ref{tab:cross_modal_datasets}.

\paragraph{Standardized Single-Task Benchmarks} To benchmark RetSAM against specialized models, we compile a suite of 17 public datasets \cite{fraz2012ensemble,staal2004ridge,jin2022fives,budai2013robust,zhang2016robust,hoover2000locating,carmona2008identification,sivaswamy2014drishti,bajwa2020g1020,wu2023gamma,zhang2010origa,kovalyk2022papila,orlando2020refuge,li2019diagnostic,decenciere2013teleophta,zhou2020benchmark,porwal2018indian} mapped to three downstream tracks. This collection includes six datasets for vessel segmentation, seven datasets for OD/OC segmentation, and four datasets for lesion segmentation. We adhere to official train/test splits where available. For datasets lacking official partitions, we establish a standardized random split of 70\% training, 10\% validation, and 20\% testing. This fixed partition is used across all comparative experiments to ensure fair benchmarking.

\paragraph{Unified Multi-Task Benchmarks.} Existing benchmarks are often single-task, where datasets labeled for one target ignore co-existing features. To evaluate RetSAM's ability to simultaneously perceive multiple targets, we use the MAPLES-DR dataset and construct four re-annotated benchmarks. MAPLES-DR provides dense pixel-wise annotations for 10 anatomical and pathological biomarkers across 198 images. We further extend four classic datasets, including DRIVE, HRF, GAMMA, and IDRiD, by adding missing annotations across vessel, OD/OC, and lesion domains. Regarding the lesion domain, our re-annotation focuses on hemorrhages and exudates, which are among the most prevalent findings previously unannotated in these datasets. This process is conducted by three ophthalmologists under a standardized protocol, with results validated by a senior expert. These enriched annotations will be released to facilitate future multi-task research.

\paragraph{Cross-Modal Adaptation Benchmarks.} To assess transferability to unseen modalities, we select two modalities, including UWF and OCTA to establish a testbed for domain adaptation. For UWF, we use the Prime-FP20 benchmark for vessel segmentation and a proprietary dataset of 277 high-resolution images for OD/OC segmentation. The latter is curated by the Airdoc Medical Department, with annotations performed by a senior ophthalmologist adhering to clinical standards. For OCTA, we use the OCTA-500 dataset, targeting artery and vein separation in 2D projection maps. To evaluate robustness under data scarcity, we adopt a strict few-shot protocol: 5\% of the data is allocated for training and 2\% for validation, with all remaining data reserved for testing. This setting evaluates the model's ability to adapt to novel domains with minimal supervision.

\subsection*{Network Architecture}

RetSAM is a fully supervised framework for multi-target segmentation in retinal imaging. Built on SwinUNETR, RetSAM uses a hierarchical Swin Transformer encoder to capture both local detail and global context. Unlike standard architectures that use a single decoder for all targets, RetSAM adopts a single-encoder multi-decoder design to decouple learning across tasks.

\paragraph{Hierarchical Visual Encoder} We use a Swin-ViT-base backbone and adjust its hyperparameters for high-resolution fundus analysis. The network processes input images at a resolution of $640 \times 640$. The encoder uses a patch size of 4, a window size of 10, and an embedding dimension of $C=128$. The Swin Transformer blocks have depths of $[2, 2, 18, 2]$ across four stages, with attention heads of $[4, 8, 16, 32]$. Multi-scale feature maps are passed to the decoders via skip connections. This design also allows future adjustments to input resolution and model capacity to accommodate different clinical environments. Detailed specifications are provided in Table~\ref{tab:implementation_details}.

\paragraph{Task-Decoupled Multi-Decoder} RetSAM separates decoding paths across tasks. Standard architectures often use one decoder for all classes, which can introduce optimization conflicts when targets differ substantially in shape and appearance. RetSAM mitigates this by assigning independent decoder branches to specific tasks. Each decoder follows a U-Net-like upsampling structure to generate outputs independently. The output channels of each decoder match the number of categories within its task, reducing cross-task interference during multi-class prediction.

\subsection*{Deterministic Quantitative Analysis}

Following segmentation, we apply a deterministic post-processing pipeline to convert predicted masks into clinical biomarkers. The pipeline uses rule-based image analysis procedures to promote reproducibility. The analysis includes the following components.

\paragraph{Vascular Topology and Morphology} To quantify retinal vessels, we skeletonize vessel masks to obtain single-pixel centerlines. We estimate vessel caliber using the Euclidean distance transform on the vessel mask. Following clinical conventions, measurements are computed within a standard region of interest defined as the annulus between 1.5 and 2.0 disc radii from the optic disc center. We also quantify geometric complexity using box-counting fractal dimension and curvature-based tortuosity.

\paragraph{Optic Disc and Cup Geometry} For optic disc analysis, we smooth the raw segmentation masks using a convex hull operation. The system determines laterality of the eye to orient the ISNT (Inferior, Superior, Nasal, Temporal) neuroretinal rim analysis, ensuring anatomically consistent measurements.

\paragraph{Pathological Burden and Distribution} For lesion analysis, we compute burden metrics using connected-component analysis, including global indicators such as total area and density, as well as instance-level descriptors such as circularity. To characterize spatial distribution, the framework divides the fundus into quadrants centered on the macula and adjusts nasal/temporal orientation based on eye laterality.

\paragraph{} By applying these procedures across RetSAM tasks, we generate a total of 30 quantitative metrics. A summary of these indicators is provided in Table~\ref{tab:feature_definitions}.

\subsection*{Training Strategy}

We adopt a three-stage training strategy to balance specialist accuracy, broad task coverage, and robustness across acquisition settings. The overall goal is to train a single model in the unified RetSAM task space while leveraging high-quality private annotations and the scale and diversity of public fundus collections.

\paragraph{Stage 1: Task-Specific Expert Training}
We first train a set of task-specific expert models on high-quality private datasets. Each expert uses the same baseline architecture as RetSAM and predicts labels in the unified RetSAM task space, but is trained using task-relevant supervision only, such as vessel, optic disc and cup, or lesion segmentation. All parameters are updated during this stage. We train each expert to its best performance on the corresponding private validation data before it is used for pseudo-label generation, aiming to provide reliable teacher predictions. We use task-specific training objectives during expert learning, including topology-aware losses for vessel segmentation and focal Tversky losses for lesion segmentation to better handle small and highly imbalanced lesion regions.

\paragraph{Stage 2: Multi-Task Pre-training via Pseudo-Labeling}
We then use the Stage 1 experts to generate pseudo-labels for large-scale public fundus images, converting public data into a unified multi-task training set aligned with the RetSAM label definitions. To control pseudo-label noise, we apply probability thresholding on the softmax outputs and retain only pixels with confidence greater than 0.75, producing hard masks after thresholding. For tasks with strong topological requirements, such as vessel segmentation, we additionally perform image-level filtering and exclude samples with poor predicted topology, including severely disconnected masks, excessive fragmentation, or an abnormal number of short spurious branches. Pseudo-label generation uses the same input resolution and preprocessing as model training. We then train a single RetSAM model on the resulting pseudo-labeled multi-task data, enabling it to learn shared representations across targets while being exposed to diverse acquisition conditions.

\paragraph{Stage 3: Task-Specific Adaptation}
Finally, we adapt the Stage 2 model to the private datasets to align predictions with the annotation standards used for deployment and downstream biomarker extraction. In this stage, we freeze the encoder to preserve general representations learned from large-scale public data and fine-tune only the decoders. This design helps limit forgetting of features learned in Stage 2 and reduces task interference, while allowing task heads to adjust their outputs to the private label conventions. The resulting model therefore retains broad robustness while producing task-consistent segmentations suitable for standardized quantification.

\subsection*{Training Implementation}

We implement the training protocols in PyTorch using the MONAI library, adapting configurations to the requirements of each stage. To improve robustness, we apply online augmentations throughout all stages, including random rotations, flips, color jitter, and grayscale conversion. For single-task expert training (Stage 1) and Stage 3 fine-tuning, we use AdamW with an initial learning rate of $1 \times 10^{-4}$ and a cosine-annealing schedule over 100 epochs. We train with a batch size of 16 and optimize a hybrid objective that combines Dice loss and cross-entropy loss with weights 1.5 and 1.0, respectively.

In contrast, large-scale pre-training of the unified RetSAM model (Stage 2) uses a different configuration to promote stable convergence on the pseudo-labeled dataset, which comprises approximately 160{,}000 images and 1.2 million pseudo-labeled masks across tasks. For this stage, we reduce the initial learning rate to $1 \times 10^{-5}$, extend training to 400 epochs, and use a batch size of 4 per GPU. To balance multi-task learning, we aggregate task-specific losses via a weighted summation with task-specific coefficients. The resulting weights are used to initialize the subsequent fine-tuning stage. Detailed training configurations and hyperparameters are provided in Table~\ref{tab:implementation_details}.

\subsection*{Computational Resources}

All experiments are conducted on a server equipped with 8 NVIDIA A800 GPUs, each with 80~GB of memory. We allocate resources based on model scale. Each single-task baseline and fine-tuned model is trained on a single GPU, requiring approximately 32 GPU-hours per model. Pre-training the core RetSAM model uses multi-GPU acceleration with PyTorch DistributedDataParallel and requires approximately 416 GPU-hours. To support reproducibility, we standardize the software environment using Python 3.10.12, PyTorch 2.6.0, PyTorch Lightning 2.6.0, and MONAI 1.4.0. For comparative evaluations, we follow the official codebases and configurations of representative methods, including MedSAM, MedSAM2, Medical-SAM-Adapter, SAM2, SAM3, SAM2-UNet, and RetFound. We use identical train--test splits across experiments to ensure fair comparison.

\subsection*{Experimental Setting}

\paragraph{Baseline Implementation.} To ensure fair comparison, we run all compared methods, including SAM-based models, multi-task specialists, and self-supervised pre-trained models, using their official repositories. We adhere to the hyperparameter settings provided in the corresponding codebases whenever applicable. All models are trained and evaluated on the same data splits as RetSAM to ensure consistent benchmarking.

\paragraph{Evaluation Strategy for RetSAM.} We evaluate RetSAM under two protocols: zero-shot linear inference and task-specific fine-tuning. In the zero-shot linear inference setting, we apply RetSAM to the test images without any weight updates to generate segmentation outputs. In the task-specific fine-tuning setting, we initialize from the RetSAM checkpoint and fine-tune on each task using the full training set. Unless otherwise specified, we fine-tune for 50 epochs with AdamW using a learning rate of $1 \times 10^{-4}$.

\paragraph{Cross-Modality Generalization.} To assess robustness under unseen acquisition protocols, we use the UWF and OCTA datasets described in the previous section. We adopt full-parameter fine-tuning and update all model parameters on the target modality. We fine-tune for 50 epochs using the same optimizer and learning-rate schedule as in the primary downstream evaluations.

\paragraph{Evaluation Metrics.} We report metrics in three categories.

\textbf{Region consistency.} We use DSC and JAC as primary measures of segmentation overlap across tasks.

\textbf{Boundary and topology.} We report the 95 percentile Hausdorff distance (HD95) for optic disc and cup segmentation. For vessel segmentation, we report centerline Dice (clDice) to quantify structural connectivity.

\textbf{Pixel-level classification.} For lesion segmentation, we additionally report precision to assess pixel-level detection performance.




\section*{Data Availability}

The datasets used for training the pseudo-label generating models are not publicly available due to patient privacy and ethical restrictions. In contrast, all public datasets employed for training the core RetSAM model are fully accessible. Details regarding the sources and access methods for these public datasets are listed in Table~\ref{tab:public_datasets}. The pseudo-labels generated from our private data are available upon reasonable request. Researchers interested in accessing these pseudo-labels may direct their requests to the corresponding author for assessment under appropriate data sharing agreements. 

\section*{Code Availability}

The code for RetSAM is available at \url{https://github.com/Wzhjerry/RetSAM}. For the implementation of other methods, the code can be accessed at their respective public repositories: 

\begin{itemize}
    \item \textbf{MedSAM} at \url{https://github.com/bowang-lab/MedSAM},
    \item \textbf{MedSAM-2} at \url{https://github.com/SuperMedIntel/Medical-SAM2},
    \item \textbf{Medical-SAM-Adapter} at \url{https://github.com/SuperMedIntel/Medical-SAM-Adapter},
    \item \textbf{SAM2} at \url{https://github.com/facebookresearch/sam2},
    \item \textbf{SAM3} at \url{https://github.com/facebookresearch/sam3}
    \item \textbf{SAM2-Unet} at \url{https://github.com/WZH0120/SAM2-UNet},
    \item \textbf{RetFound} at \url{https://github.com/rmaphoh/RETFound_MAE}.
\end{itemize}

\section*{Acknowledgements}

We thank the Google Cloud Platform team for providing the cloud computing services that enabled this project. We thank all the ophthalmologists and medical professionals who participated in this paper; their time and expertise were invaluable to the validation of this work. Furthermore, we thank the domain experts at Airdoc for their efforts in providing the high-quality annotated data used in this study.

\clearpage

\end{document}